%% file: main.tex
\documentclass[twoside,11pt]{article}

\usepackage{jmlr2e}
\usepackage[utf8]{inputenc} 
\usepackage[T1]{fontenc}    
\usepackage{hyperref}       
\usepackage{url}            
\usepackage{booktabs}       
\usepackage{amsfonts}       
\usepackage{nicefrac}       
\usepackage{microtype}      
\usepackage{microtype}
\usepackage{graphicx}
\usepackage{subfigure}
\usepackage{xspace}
\usepackage{amsmath}
\usepackage{enumitem}
\usepackage{booktabs} 
\usepackage{caption}
\usepackage{xcolor,ulem,comment,multirow}
\usepackage{wrapfig}

\ShortHeadings{}{}
\firstpageno{1}

\begin{document}

\title{Beyond English-Centric Multilingual Machine Translation}
\author{\centering Angela Fan\thanks{Corresponding Author. Email: \texttt{angelafan@fb.com}}, Shruti Bhosale, Holger Schwenk, Zhiyi Ma, Ahmed El-Kishky, Siddharth Goyal, Mandeep Baines, Onur Celebi, Guillaume Wenzek, Vishrav Chaudhary, Naman Goyal, Tom Birch, Vitaliy Liptchinsky, Sergey Edunov, Edouard Grave, Michael Auli$^\dagger$, Armand Joulin\thanks{Equal contribution. Order determined with coin flip.} \thanks{Holger Schwenk, Onur Celebi, Vishrav Chaudhary, Ahmed El-Kishky, Angela Fan, Edouard Grave, Zhiyi Ma, and Guillaume Wenzek worked on large-scale data mining, including improvements to fasttext, LASER, CCAligned, and CCMatrix. Siddharth Goyal worked on backtranslation. Michael Auli, Mandeep Baines, Shruti Bhosale, Tom Birch, Sergey Edunov, Angela Fan, Naman Goyal, Siddharth Goyal, and Vitaliy Liptchinsky worked on model scaling and scaling infrastructure. Michael Auli, Shruti Bhosale, Sergey Edunov, Angela Fan, Edouard Grave, Armand Joulin, Zhiyi Ma, and Holger Schwenk worked on model and experimental design. Michael Auli, Ahmed El-Kishky, Angela Fan, Edouard Grave, Armand Joulin, Zhiyi Ma, and Holger Schwenk wrote the paper.} \\ 
{\normalfont Facebook AI, $^*$LORIA}\\}
\editor{}

\maketitle

\begin{abstract}
Existing work in translation demonstrated the potential of massively multilingual machine translation by training a single model able to translate between any pair of languages. 
However, much of this work is English-Centric by training only on data which was translated from or to English.
While this is supported by large sources of training data, it does not reflect translation needs worldwide. 
In this work, we create a true Many-to-Many multilingual translation model that can translate directly between any pair of 100 languages. 
We build and open source a training dataset that covers thousands of language directions with supervised data, created through large-scale mining.
Then, we explore how to effectively increase model capacity through a combination of dense scaling and language-specific sparse parameters to create high quality models. 
Our focus on non-English-Centric models brings gains of more than 10 BLEU when directly translating between non-English directions while performing competitively to the best single systems of WMT. We open-source our scripts so that others may reproduce the data, evaluation, and final M2M-100 model \href{https://github.com/pytorch/fairseq/tree/master/examples/m2m_100}{\texttt{here}}.
\end{abstract}


\section{Introduction}
\input{intro.tex}

\input{background.tex}

\input{data.tex}

\input{comp.tex}

\input{model.tex}

\section{Bringing it all Together}
\label{sec:combine}

We have explored the creation of a true many-to-many dataset for the multilingual translation of 100 languages, as well as how to effectively scale Many-to-Many models through a mix of dense and sparse scaling. In this section, we summarize our final results, compare to existing published work --- both multilingual benchmarks and competitive directions in WMT --- and end with a human evaluation of the overall quality of our translation quality.

\subsection{Real-world Settings for Many-to-Many Translation}


We highlight that there are several real-world usecases of translation directions not involving English. For example, many countries have official and regional languages that are not English, which would be natural candidates for direct translation. For example, it is intuitive to translate Kazakh directly to Russian in Kazakhstan. In Table~\ref{tab:languages_by_country}, we compare English-Centric models to Many-to-Many on a variety of different non-English directions. We see that across the board, our M2M-100 model has drastically better performance and on average improves over 7 BLEU across these directions.


\begin{table}
\setlength{\tabcolsep}{5.5pt}
\centering
\small
    \begin{tabular}{l l l | l | c c l }
    \toprule
    & \bf Source & \bf Target & \bf Test Set & \multicolumn{3}{c}{\bf BLEU} \\
    &           &            && English-Centric & M2M-100 &  \multicolumn{1}{c}{$\Delta$} \\ 
    \midrule 
    India & Hindi & Bengali & TED & 3.9  & 8.7 & +4.8 \\ 
          & Hindi & Marathi  & TED & 0.4 & 8.4 & +8.0 \\ 
          & Hindi & Tamil  & TED & 1.1 & 7.5   & +6.4 \\ 
    \midrule 
    South Africa & Afrikaans & Xhosa & Autshumato & 0.1 & 3.6 & +3.5 \\ 
         & Afrikaans & Zulu & Autshumato & 0.3 & 3.6 & +3.3 \\ 
         & Afrikaans & Sesotho & Autshumato & 0.0 & 2.1 & +2.1 \\ 
         & Xhosa & Zulu & Autshumato & 0.1 & 3.6 & +3.5\\
         & Sesotho & Zulu & Autshumato & 0.1 &  1.2 & +1.1 \\ 
    Chad & Arabic & French & TED & 5.3 & 20.8 & +15.5 \\ 
    DR Congo & French & Swahili & Tatoeba & 1.8 & 5.7 & +3.9 \\ 
    \midrule 
    Kazakhstan & Kazakh & Russian & TED & 0.5 & 4.5 & +4.0 \\ 
    Singapore & Chinese & Tamil & TED & 0.2 & 8.0 & +7.8 \\
    \midrule 
    Austria & German & Croatian & TED  & 9.6 & 21.3 & +11.7 \\ 
            & German & Hungarian & TED & 11.3  & 17.4 & +6.1 \\  
    Belgium & Dutch & French & TED & 16.4 & 25.8 & +9.4 \\  
            & Dutch & German & TED & 18.1 & 26.3 & +8.2 \\ 
    Belarus & Belarusian & Russian & TED & 10.0 & 18.5 & +8.5 \\
    Croatia & Croatian & Serbian & TED & 22.4 & 29.8 & +7.4 \\  
            & Croatian & Hungarian & TED & 12.4 & 17.5 & +5.1 \\ 
            & Croatian & Czech &TED  & 15.2 & 22.5 & +7.3 \\ 
            & Croatian & Slovak & TED & 13.8 & 24.6 & +10.8 \\  
    Cyprus & Greek & Turkish & TED & 4.8 & 12.6 & +7.8 \\  
    Czechia & Czech & Slovak & TED & 9.5 & 28.1 & +18.6 \\  
    Finland & Finnish & Swedish & TED & 7.9 & 19.2 & +11.3 \\  
    Italy & Italian & French & TED & 18.9 & 28.8 & +9.9 \\ 
          & Italian & German & TED & 18.4 & 25.6 & +7.2 \\ 
    Moldova & Romanian & Russian & TED & 8.0 & 19.0 & +11.0 \\ 
            & Romanian & Ukrainian & TED & 8.7  & 17.3 & +8.6 \\  
    Montenegro & Albanian & Croatian & TED & 3.0  & 20.7 & +17.7 \\  
              & Albanian & Serbian &  TED & 7.8 & 20.6 & +12.8 \\ 
    Romania & Romanian & German & TED & 15.0 & 24.7 & +9.7 \\
            & Romanian & Hungarian & TED & 11.0  & 16.3 & +4.3 \\
            & Romanian & Turkish & TED & 5.1 & 12.0 & +6.9  \\
            & Romanian & Armenian & TED & 0.4 &  8.2 & +7.8 \\
    Russia & Bashkir & Russian & Tatoeba & 0.1 & 4.3 & +4.2 \\
            & Ukrainian & Russian & TED & 18.0 & 23.7 & +5.7 \\
\midrule
\multicolumn{3}{c}{} & Average & 8.0 & 15.6 & \bf +7.6 \\
    \bottomrule 
    \end{tabular}
    \caption{\textbf{Performance translating between official and official regional languages} of several nations, focusing on non-English directions.} 
    \label{tab:languages_by_country}
\end{table}

\subsection{Comparison on Various Translation Benchmarks}

Next, we compare our M2M-100 model to various existing work on different benchmarks. While the training data is not the same, we conduct this comparison to provide a reference point for the overall strength of our model. An important note is that for each of these benchmarks, there are various different tokenizers used which affect BLEU --- we follow the tokenization and BLEU calculation of each of these benchmarks, rather than the evaluation methodology of our previous results. Thus, the numbers in this subsection are not comparable to the rest of the paper, as they use the tokenization of each benchmark. Further, this comparison was prepared in advance, so all sentences appearing in these evaluation sets were removed from the training data we used.

\paragraph{Comparison on WMT.} First, we compare our Many-to-Many model to submissions to WMT, the premier translation competition. We display results on a variety of different language directions, some of which are standard natural language processing machine translation benchmarks, such as English-French, English-German, and English-Russian. Results are shown in Table~\ref{tab:m2m_wmt}.
\footnote{
    $\textbf{En} \leftrightarrow{} \textbf{De/En} \leftrightarrow{} \textbf{Ru:}$ we evaluated publicly available single model checkpoints prior to finetuning from~\citet{ng2019facebook} on WMT2019.
    $\textbf{En} \leftrightarrow{} \textbf{Zh:}$ we report results from~\citet{li2019niutrans} which contains single model BLEU results on WMT2019.
    $\textbf{En} \leftrightarrow{} \textbf{Lt:}$ we report results from~\citet{pinnis-etal-2019-tildes} on WMT2019; both directions are the best single model systems which use unconstrained training data. 
    $\textbf{En} \rightarrow{} \textbf{Fr:}$ we report results from~\citet{edunov2018bt}. $\textbf{Fr} \rightarrow{} \textbf{En:}$ we report results from~\citet{johnson2017google} on WMT2014. 
    $\textbf{En} \leftrightarrow{} \textbf{Lv:}$ we report results from~\citet{pinnis-etal-2017-tildes} on WMT2017. 
    $\textbf{En} \leftrightarrow{} \textbf{Tr:}$ we report results from~\citet{sennrich-etal-2017-university} on WMT17. 
    $\textbf{En} \leftrightarrow{} \textbf{Et:}$ we report results from~\citet{pinnis-etal-2018-tildes} on WMT18. 
    $\textbf{En} \leftrightarrow{} \textbf{Fi:}$ we report results from~\citet{talman-etal-2019-university} on WMT17.
    } 
Many submissions to the WMT shared task use ensembling, in-domain finetuning, or reranking methods, which are standard techniques to improve quality. As these could be added to our system at inference time as well, we focus instead on comparing single model results. To identify comparisons, we examine the WMT Shared Task proceedings as well as the submissions at \url{http://matrix.statmt.org/}. 

As seen in Table~\ref{tab:m2m_wmt}, our M2M-100 system can achieve very competitive performance compared to bilingual models tuned especially for individual WMT translation directions.
This shows that our model maintains strong translation quality on individual directions.

\begin{table}[t]
    \setlength{\tabcolsep}{5.5pt}
    \centering
    \footnotesize
        \begin{tabular}{c l|l| cccc}
        \toprule
         &  & & \multicolumn{3}{c}{\bf BLEU} \\
        \multicolumn{2}{l|}{\bf Direction} & \bf Test Set & Published  &  M2M-100 & \bf $\Delta$\\
        \midrule
        \multicolumn{2}{l}{\bf Without Improvement} \\
        & English-Chinese \citep{li2019niutrans} & WMT'19 &  38.2 & 33.2 & -5.0 \\
        & English-Finnish \citep{talman-etal-2019-university} & WMT'17 &  28.6 & 28.2 & -0.4 \\
        & English-Estonian \citep{pinnis-etal-2018-tildes} & WMT'18 & 24.4 & 24.1 & -0.3\\
        & Chinese-English \citep{li2019niutrans} & WMT'19 &  29.1 & 29.0 & -0.1\\

        \midrule 
        \multicolumn{2}{l}{\bf With Improvement} \\ 
        & English-French \citep{edunov2018bt} & WMT'14  & 43.8 & 43.8 & 0 \\
        & English-Latvian \citep{pinnis-etal-2017-tildes} & WMT'17 & 20.0 & 20.5 & +0.5 \\
        & German-English \citep{ng2019facebook} & WMT'19 &  39.2 & 40.1 & +0.9 \\
        & Lithuanian-English \citep{pinnis-etal-2019-tildes}~~~~~ & WMT'19  & 31.7 & 32.9 & +1.2\\
        & English-Russian \citep{ng2019facebook} & WMT'19 &  31.9 & 33.3 & +1.4\\
        & English-Lithuanian \citep{pinnis-etal-2019-tildes} & WMT'19 & 19.1 & 20.7 & +1.6 \\
        & Finnish-English \citep{talman-etal-2019-university} & WMT'17 & 32.7 & 34.3 & +1.6 \\
        & Estonian-English \citep{pinnis-etal-2018-tildes} & WMT'18 & 30.9 & 33.4 & +2.5 \\
        & Latvian-English \citep{pinnis-etal-2017-tildes} & WMT'17 & 21.9 & 24.5 & +2.6 \\
        & Russian-English \citep{ng2019facebook} & WMT'19 &  37.2 & 40.5 & +3.3 \\
        & French-English \citep{edunov2018bt} & WMT'14  & 36.8 & 40.4 & +3.6 \\
        & English-German \citep{ng2019facebook} & WMT'19 &  38.1 & 43.2 & +5.1 \\
        & English-Turkish \citep{sennrich-etal-2017-university} & WMT'17 & 16.2 & 23.7 & +7.5 \\
        & Turkish-English \citep{sennrich-etal-2017-university} & WMT'17 & 20.6 & 28.2 & +7.6 \\
        \midrule 
        \multicolumn{3}{r}{Average} & 30.0 & 31.9 & \bf +1.9  \\ 
        \bottomrule
        \end{tabular}
    \caption{\textbf{Comparison of Many-to-Many and public results on WMT datasets.} We compare M2M-100 to published work (best single models) on WMT. 
    To identify previous work, we examine the WMT Shared Task proceedings for the top performing models and check reported results on \url{http://matrix.statmt.org/}. 
    For these comparisons, we report detokenized BLEU with sacrebleu~\citep{post2018sacrebleu} on the test set.}
    \label{tab:m2m_wmt}
\end{table}


Next, we compare our models to other multilingual translation work. Table~\ref{tab:comparison_results} displays several previously published results on different sets of benchmarks. Note that for each comparison, we follow the published setting in tokenization, evaluation, and whether or not the BLEU is tuned on the validation set to maximize comparability. 


\paragraph{Bilingual Models.} We first compare to mBART~\citep{liu2020multilingual}, which creates bilingual models based on finetuning a pretrained model on individual language directions. After pretraining as a denoising autoencoder, publicly available bitext data is used to create various different bilingual models, one for each evaluation direction. \citet{liu2020multilingual} tune the test set BLEU on the validation set. Following their setting, we tune the generation beam size between \{5,10\}, and length penalty between \{0.5, 1.0, 1.5\}, and the number of checkpoints to average between \{1, 5, 10\}. Our model provides +0.7 BLEU improvement.

We then compare to the bilingual baselines provided in CCMatrix~\citep{schwenk2019ccmatrix}, which trained individual models for each direction. As these models generate with no tuning, we generate on all pairs with beam size 5 and length penalty 1, using only the best checkpoint. Our one Many-to-Many multilingual model achieves a 2 BLEU point gain on average compared to training hundreds of individual models. 

\paragraph{Multilingual Models.} We next compare the performance of our multilingual system to other published multilingual systems. We compare to the English-Centric multilingual model from~\citet{zhang2020improving} on the OPUS100 corpus. Their model is trained with noisily aligned through-English data from OPUS~\citep{tiedemann2012opus,zhang2020improving}, with online backtranslation to improve the performance of non-English pairs. Note that \citet{zhang2020improving} train on 100 directions, but we only overlap a subset of directions. However, we fully cover their full set of non-English evaluation pairs. Finally, the OPUS100 non-English directions come only with a test set, so we generate with beam size 5, length penalty 1, and use the best checkpoint. As shown in Table~\ref{tab:comparison_results}, we improve by more than 4 BLEU.

\begin{table}[t]
\setlength{\tabcolsep}{5.5pt}
\centering
\small 
    \begin{tabular}{ll cc }
    \toprule
    \bf Benchmark & \bf Model && \bf BLEU \\
    \midrule 
    \multirow{2}{*}{\textbf{mBART}} & Previous Work~\citep{liu2020multilingual} && 23.9 \\ 
       & M2M-100 &&  \bf 24.6 \\ 
    \midrule 
    \multirow{2}{*}{\textbf{CCMatrix}} & Previous Work~\citep{schwenk2019ccmatrix} && 16.3 \\
       & M2M-100 && \bf 18.7\\ 
    \midrule 
    \multirow{2}{*}{\textbf{OPUS100}} & Previous Work~\citep{zhang2020improving} && 14.1 \\ 
       & M2M-100 && \bf 18.4 \\ 
    \bottomrule
    \end{tabular}
    \caption{\textbf{Comparison on various evaluation settings from previous work}. We display the best performing model from the published work and report average BLEU on the test set. For these comparisons, we use the tokenization and BLEU evaluation script used by each work for comparability. \citet{liu2020multilingual} report Low/Mid resource directions into and out of English and High resource directions into English, we average across all. \citet{schwenk2019ccmatrix} report the full matrix on 28 languages, we average across all. \citet{zhang2020improving} report results on non-English directions, we average across all.} 
    \label{tab:comparison_results}
\end{table}

\subsection{Human Evaluation}

\begin{figure}[t]
    \centering
    \includegraphics[width=\textwidth]{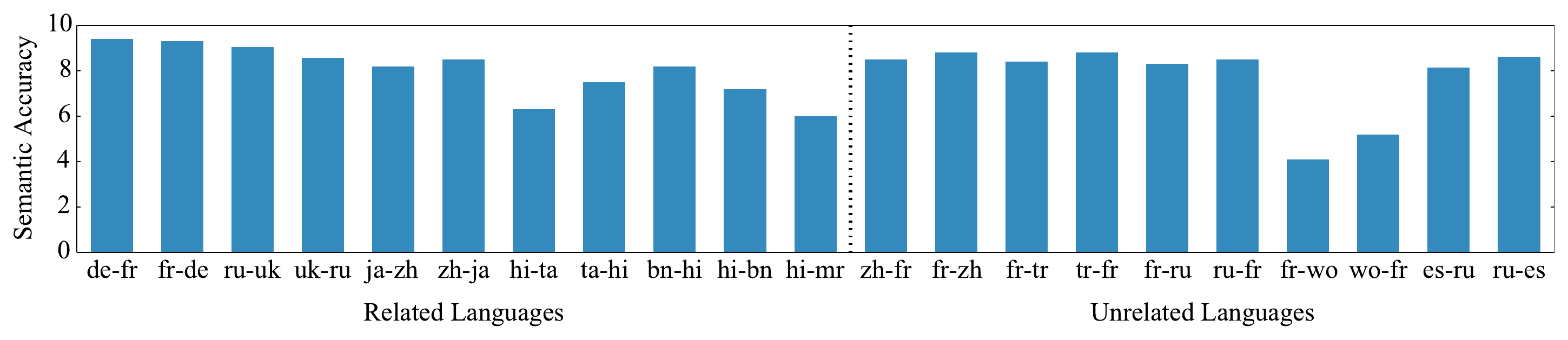} 
    \caption{
    \textbf{Human Evaluation of Translation Accuracy of M2M-100 on Non-English Directions.} Evaluators are asked to score the semantic accuracy of translations on a scale of 1 to 10. 
    }
    \label{fig:human_eval}
\end{figure}

\begin{figure}[t]
    \centering
    \includegraphics[width=0.98\textwidth]{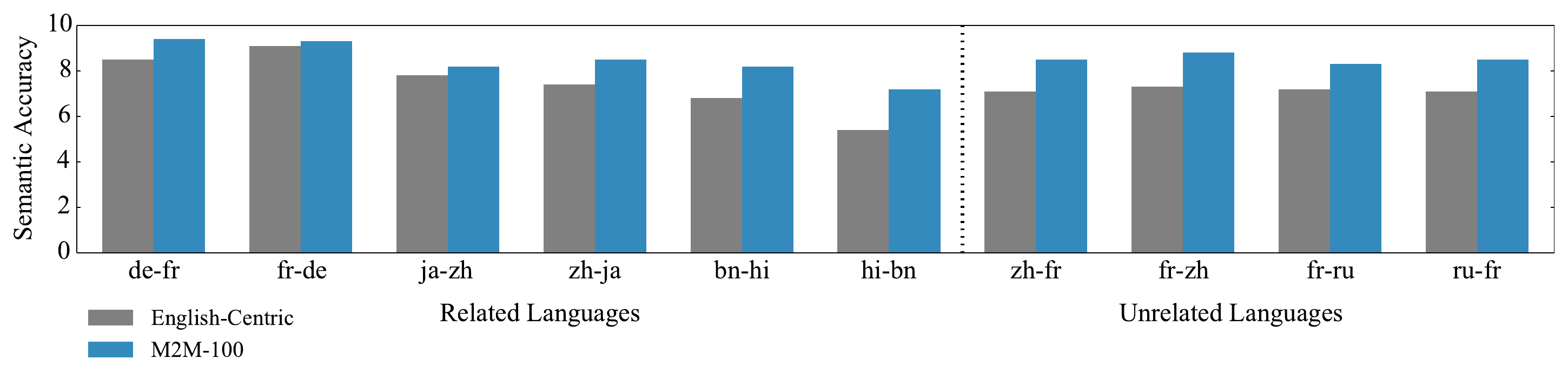} 
    \caption{
    \textbf{Human Evaluation of Translation Accuracy of M2M-100 compared to English-Centric on 10 Non-English Directions.} Evaluators are asked to score the semantic accuracy of translations on a scale of 1 to 10. 
    }
    \label{fig:human_eval_2}
\end{figure}

We end with a human evaluation study to understand the quality of our model translations. We focus on 20 different directions, none of them involving English. We include languages commonly spoken in the same region, such as Japanese-Chinese, Hindi-Tamil, and Russian-Ukrainian, as well as directions that cross language families, such as Chinese-French, French-Arabic, and Russian-Spanish. We also include several very low resource directions, such as French-Wolof, Hindi-Marathi, and Japanese-Mongolian. All of our evaluators are native speakers in one of the languages and fluent in the other.

Each evaluator rates 50 different translations for semantic accuracy on a scale of 1 to 10. Results are shown in Figure~\ref{fig:human_eval}. On semantic accuracy, most of our evaluations score between 8.5 and 9.5 (with 10 being the best possible score). For lower resource directions, the scores remain reasonable. Hindi to Tamil and Wolof to French score around 7-8. The most challenging direction based on human evaluation is French into Wolof (fr-wo), likely because there is not sufficient target-side Wolof data. 

Next, we compare our model with an English-Centric model on 10 directions in Figure~\ref{fig:human_eval_2}. Each evaluator is asked to rate 100 sentences, 50 from each model, in a blind test. Across the board, we find that our Many-to-Many system scores better in translation accuracy - both for related and unrelated languages.

\subsection{Discussion}

\paragraph{Curating High-Quality Training Data.} 
Creating high quality datasets to train translation models has been a long-standing area of research. For example, previous work has explored how to best filter noisy datasets~\citep{koehn2018findings,koehn2019findings}. Our use of large-scale mined training data presents large quantities of data to train multilingual models on, but brings challenges as well. For example, our mining methods mine both simplified and traditional Chinese text, tokenized and untokenized text, and many examples with code switching. We apply several data filtering methods, but the cleanliness and quality of alignment is critical for training high-quality translation systems. Further, multilingual translation can be affected by domain mismatch, as people in different parts of the world discuss different topics~\citep{shen2019source}, which presents additional challenges for curating good training sets. Thus, we see the continued improvement of data quality as an important direction for multilingual translation systems, which require a lot of data to train well.

\paragraph{Improvements on Very Low-Resource Languages.} 
Strong performance for low-resource languages remains a critical area for future improvement~\citep{gu2018universal,sennrich-zhang-2019-revisiting}. For many languages, our system still requires substantial improvements. Examples include African languages such as Xhosa and Zulu, European languages such as Catalan and Basque, and Southeast Asian languages such as Iloko and Cebuano. For many of these, even monolingual resources on the internet are limited, which strongly affects the quantity and quality of mined data. Using curated data, possibly supplemented by mining, may provide a starting point for future improvement. For example, several resources for African languages exist, including JW300~\citep{agic-vulic-2019-jw300} used in the \textsc{masakhane} machine translation effort~\citep{orife2020masakhane} and datasets for Nigerian Pidgin~\citep{ahia2020towards}, Wolof~\citep{alla2020using}, Fon~\citep{emezue2020ffr}, Igbo~\citep{ezeani2020igbo}, Amharic, Tigrigna, Afan-Oromo, Wolaytta, and Ge'ez~\citep{abate2018parallel}. Other lines of work present resources for low-resource Asian languages, such as the ALT project~\citep{riza2016introduction,ding2016similar}, Mongolian, Uyghur, and Tibetian~\citep{anonymous2020}, or strategies for improvement on specific directions~\citep{chen2019facebook}. Further research is required to bring together small datasets of higher quality translations, mined data, and monolingual resources to create improved translation systems for very low resource languages.

\section{Conclusion}

We introduced M2M-100, a new Many-to-Many multilingual translation model that can translate between the 9,900 directions of 100 languages.
The underlying dataset was mined from CommonCrawl using a novel strategy which exploits  language groupings to avoid mining every possible direction while maintaining good accuracy.
Such a large dataset requires models with increased capacity and to this end we explored densely scaling the number of parameters as well as sparsely, through introducing language-specific parameters trained with a novel random re-routing scheme.

Results show that M2M-100 outperforms English-Centric multilingual models trained on data where either the source or target language is English. 
The system improves over 10 BLEU on average compared to an English-Centric baseline when translating directly between non-English directions.
M2M-100 is competitive to bilingual models from WMT and improves over existing publicly available multilingual translation systems.
Human judges indicate that our model translates fluently with high semantic accuracy.

\input{results.tex}

\acks{We thank Yuqing Tang and Peng-Jen Chen for their work on the multilingual translation infrastructure in fairseq. We thank Xian Li, Chau Tran, Yuqing Tang, Peng-Jen Chen, and Marc'Aurelio Ranzato for insightful conversations. We thank our volunteer human evaluators for closely examining the translation quality of our models through various directions.}


\bibliography{sample}

\newpage
\appendix
\input{appendix.tex}

\end{document}

%% file: intro.tex
Multilingual Machine Translation~(MMT) aims to build a single model to translate between any pair of languages.
Neural network models have been very successful for bilingual machine translation~\citep{bahdanau2014neural,gehring2017convs2s,vaswani2017attention} and more recently, neural MMT models have shown promising results~\citep{firat2016multi,zhang2020improving}.
Multilingual translation models factorize computation when translating to many languages and share information between similar languages, which benefits low resource directions~\citep{arivazhagan2019massively} and enables zero-shot translation~\citep{gu2019improved}.

However, in the past, these systems have not performed as well as bilingual models when trained on the same language pairs~\citep{johnson2017google}, as model capacity necessarily must be split between many languages~\citep{arivazhagan2019massively}.
This has been alleviated by increasing model capacity~\citep{aharoni2019massively,zhang2020improving}, but increased model size also necessitates larger multilingual training datasets which are laborious and difficult to create.
To ease this challenge, most prior work has focused on \textit{English-Centric} datasets and models which translate from and to English but not between non-English languages. 
This English-Centric bias in the data and resulting models is not reflective of how people use translation and empirically leads to lower performance for non-English translation directions.

In this work, we create more diverse multilingual machine translation models by building a large-scale Many-to-Many dataset for 100 languages. 
We considerably reduce the complexity of this task through the automatic construction of parallel corpora~\citep{artetxe2018margin,schwenk2019ccmatrix} with a novel data mining strategy that exploits language similarity to avoid mining all directions.
We also leverage backtranslation to improve the quality of our model on zero-shot and low resource language pairs.
Overall, we build the first true Many-to-Many dataset comprising 7.5B training  sentences for $100$ languages, providing direct training data for thousands of translation directions.

The quantity of data in a Many-to-Many dataset increases quadratically with the number of languages, making neural networks with standard capacity underfit rapidly.
To that effect, we leverage progress in scaling~\citep{kaplan2020scaling,arora2018optimization} to train models that are over $50$ times larger than current bilingual models with model parallelism~\citep{huang2019gpipe,shoeybi2019megatron}. 
Even with these tools, scaling the number of parameters hardly follows the quadratic increase in data induced by the Many-to-Many setting, and we propose several scaling strategies tailored to the specificities of our problem.
In particular, we consider a deterministic mixture-of-experts strategy to split the model parameters into non-overlapping groups of languages which we train with a novel re-routing strategy.
Language specific mixture-of-experts also reduce the need to densely update parameters and are more parallelizable in a multi-machine setting.
Overall, combining these strategies allows us to scale the capacity of the models to a size of $15.4$B parameters and still train them efficiently on hundreds of GPUs. 

The resulting method allows us to scale Transformers and directly translate between 100 languages without pivoting through English at a performance that is competitive with bilingual models on many competitive benchmarks, including WMT.
Figure~\ref{fig:pullfigure} illustrates our data mining strategy as well as our model architecture.
This paper is organized as follows: first, we introduce several standard components of modern machine translation and explain how they apply in the multilingual setting (\autoref{sec:prelim}), then describe our strategy to scale the number of language pairs to create a Many-to-Many dataset (\autoref{sec:data}).
We then systematically compare this Many-to-Many dataset to an English-Centric approach (\autoref{sec:comp}).
Next, we incorporate increased model scale through both dense scaling and sparse mixture-of-experts (\autoref{sec:model}). Finally, we end with a thorough analysis, including human evaluation, of the quality of our 100x100 Many-to-Many translation system (\autoref{sec:combine}).

%% file: background.tex
\begin{figure}[t]
    \centering
    \includegraphics[width=\textwidth]{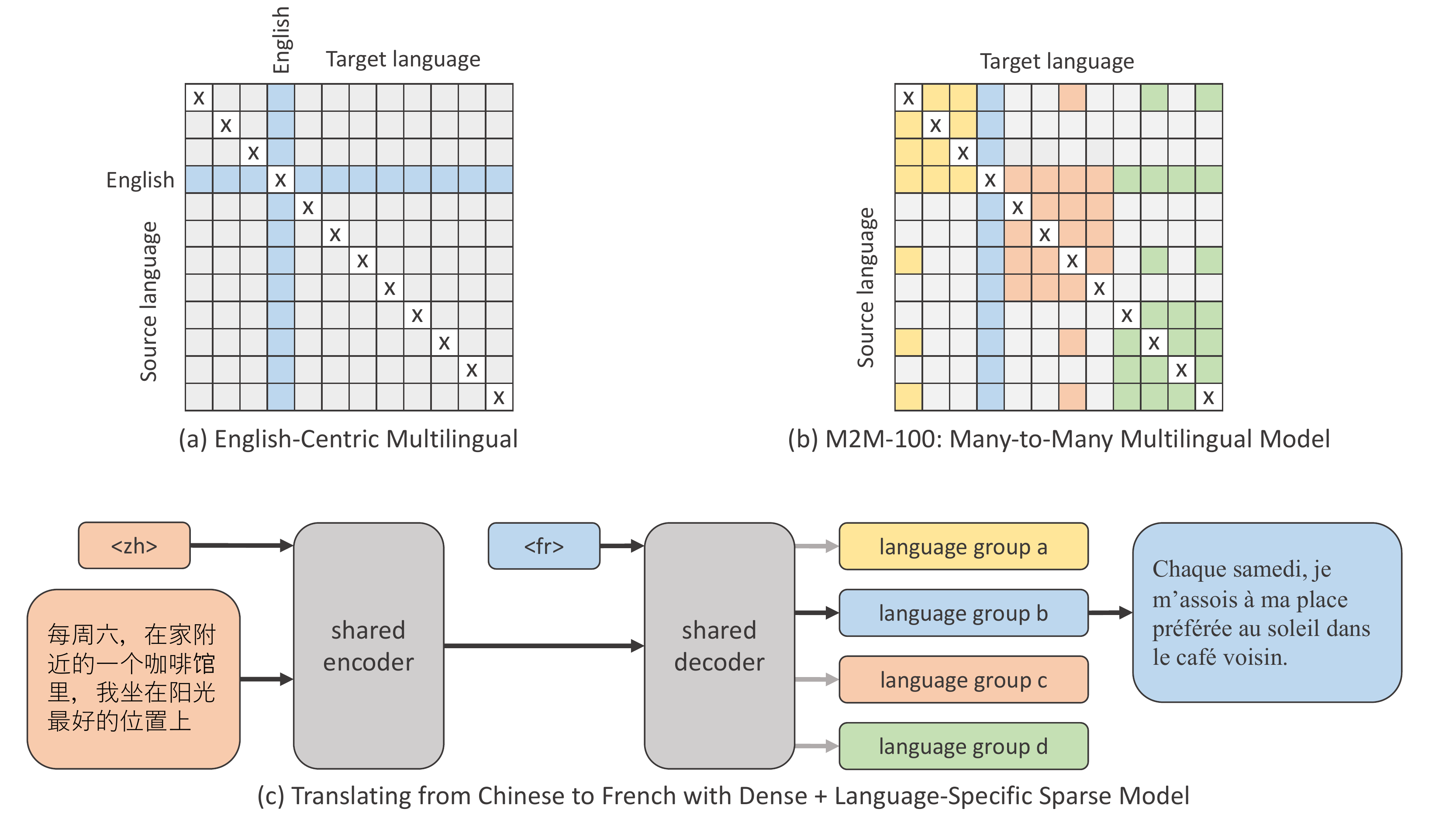} 
    \caption{\textbf{Summary of our Many-to-Many dataset and multilingual model}. English-Centric data (top left) only contains training data to and from English, where as our Many-to-Many multilingual setting (top right) contains data directly through various different directions. Our proposed model, M2M-100, combines dense and sparse language-specific parameters to translate directly between languages (bottom).} 
    \label{fig:pullfigure}
\end{figure}

\section{Preliminaries}
\label{sec:prelim}

In this work, we investigate how we can best translate from 100 languages to 100 languages, or 9900 directions, using a single model.
We describe our starting point in this section, and provide preliminary context on Transformer-based neural machine translation models.

Sequence-to-sequence models are trained on pairs of sequences, conditioning on an input sequence to produce an output sequence.
Each sentence is split into tokens, that can be words or characters, resulting in pairs of sequences $(w_1,\dots,w_S)$ and $(v_1,\dots,v_T)$.
Most machine translation systems are trained by maximizing the probability of the target sequence, given the source sentence and the target language $\ell_t$:
$$P(v_1,\ \dots,\ v_T~|~w_1,\ \dots,\ w_S,\ \ell_t)$$
Modern neural machine translation systems are based on several standard components, namely a subword segmentation method and an encoder-decoder architecture called a Transformer.
We describe these components in the context of multilingual translation.
 
\paragraph{Segmentation with SentencePiece.}
The input and output of translation systems are sequences of tokens.
These tokens are units from a dictionary built with the goal to reconstruct any sentence in any language.
Using words as base units is challenging, as it leads either to vocabularies with poor coverage or to large vocabularies.
This is especially true in the multilingual setting.
Another limitation of word-based systems are languages that are not naturally split into words, like Thai.
An alternative approach is to use \textit{subword} units, which are learned directly from  data~\citep{sennrich2015neural,kudo2018sentencepiece}.
We use SentencePiece\footnote{\url{https://github.com/google/sentencepiece}} as it was designed to work with languages with no segmentation, making it particularly suited to our setting.
We train a model with 0.9995 character coverage to have sufficient representation of character-based languages.

\paragraph{Creating a Multilingual Dictionary.}
SentencePiece produces subword units depending on their frequency in the training dataset.
Naively applying it to our corpora would result in low resource languages and languages written in less frequent scripts being underrepresented in the resulting  dictionary.
Randomly sampling data favors overrepresented languages because the probability of picking language $\ell$ is proportional to its number of sentences, $D_\ell$, i.e., $p_\ell = \frac{D_\ell}{\sum_i{D_i}}$.
We circumvent this problem by adding monolingual data for low resource languages and by using temperature sampling with $T=5$. 
More precisely, the probability $p_\ell$ is rescaled to $p_\ell^{\frac{1}{T}}$ where the temperature $T$ controls the distribution.
For example, setting $T$ to $1$ gives the original data distribution.
The resulting dictionary contains 128k unique tokens that are well distributed across languages, as shown in Appendix~\ref{app:dictionary}.

\subsection{Transformers}
Our multilingual machine translation model is based on the Transformer sequence-to-sequence architecture, which is composed of two modules: the encoder and the decoder~\citep{vaswani2017attention}.
The encoder transforms the source token sequence into a sequence of embeddings of the same length.
Then, the decoder sequentially produces the target sentence, token by token, or autoregressively.
More precisely, the encoder takes the sequence of tokens $W=(w_1,\dots,w_S)$ and the source language $\ell_s$, and produces a sequence of embeddings $H=(h_1,\dots,h_S)$, which are then fed to the decoder with the target language $\ell_t$ to produce the sequence of target tokens $V=(v_1,\dots,v_T)$ sequentially, i.e.,
\begin{eqnarray}
H &=& \texttt{encoder} (W,\ \ell_s),\\
\forall i\in[1,\dots,T],~v_{i+1} &=& \texttt{decoder} (H,\ \ell_t,\ v_1,\ \dots,\ v_i).
\end{eqnarray}
Both the encoder and decoder are composed of the same type of layers, called Transformer layers.
Each Transformer layer takes a sequence of vectors as input and outputs a sequence of vectors.
In the encoder, transformer layers are composed of two sublayers, a self-attention and a feed-forward layer.
These are applied sequentially and are both followed by a residual connection~\citep{he2015deep} and layer normalization~\citep{ba2016layer}:
\begin{eqnarray}
Z &=& \texttt{norm}\left(X + \texttt{self-attention}(X) \right),\\
Y &=& \texttt{norm}\left(Z + \texttt{feed-forward}(Z) \right).
\end{eqnarray}
The self-attention layer is an attention layer that updates each element of the sequence by looking at the other elements, while the feed-forward layer (FFN) passes each element of the sequence independently through a 2-layer MLP.
In the decoder, there is an additional third sublayer, between the self-attention and the feed-forward, which computes attention over the output of the encoder.
We refer the reader to \citet{vaswani2017attention} for details of these layers.

\paragraph{Target language token.}
The Transformer architecture has been designed for the bilingual case, where the target language is fixed.
In the case of multilingual machine translation, the target language is not fixed, and several strategies can be applied to condition the network to produce a sentence in the desired target language.
Similarly to~\citet{ha2016toward} and~\citet{johnson2017google}, we add a special token in the encoder indicating the source language and a special token in the decoder indicating the target language.

\paragraph{Training.}
Our starting point for improving massively multilingual translation models is a large Transformer model, with 12 Encoder and 12 Decoder layers, with 8192 FFN size and 1024 embedding dimension. We share the weight matrices of the input and output embeddings. The total parameter count is 1.2B. We train with the Adam optimizer~\citep{kingma2015adam} and warmup first for 4000 updates, with label smoothing $0.1$~\citep{szegedy:inception:2015,pereyra:regularize:2017}. For regularization, we tune the dropout parameter between $\{0.1, 0.2, 0.3\}$. To stabilize the training of deeper Transformers, we train with LayerDrop~\citep{fan2019reducing} 0.05 and pre-normalization~\citep{nguyen2019transformers}. 

To train with billions of sentences, we split the training data into 256 different shards to manage memory consumption. However, directly dividing mid and low resource languages into shards would reduce the variability of each shard's data for mid or low resource languages. Imagine the case where there are only $100$ sentences of a language direction per shard --- the model would easily overfit. Thus, each language is divided into a different number of shards based on resource level, such that high resource languages have more shards and the lowest resource languages only have one shard. Subsequently, lower resource shards are replicated until the full number of shards is reached.

\paragraph{Generation.}

Unless otherwise specified: for all results, we report single models with no checkpoint averaging, use beam search with beam 5, and do not tune length penalty.

%% file: data.tex
\section{Building a Many-to-Many Parallel Dataset for 100 Languages}
\label{sec:data}

In this section, we provide an overview of our Many-to-Many setting: the selection of the $100$ languages, the evaluation benchmarks, and the construction of a large-scale training set through data mining~\citep{artetxe2018margin} and backtranslation~\citep{sennrich2016backtranslation} that provides training data thousands of directions. 

\subsection{Creating a Multilingual Benchmark}

The first step of establishing a Many-to-Many dataset is to select $100$ languages for which there already exist high-quality, annotated datasets that can be used for model evaluation.

\subsubsection{Language Selection}

\newcommand{\tabtl}[1]{\begin{tabular}[h]{@{}l@{}} #1 \end{tabular}}
\newcommand{\MC}{\multicolumn}
\newcommand{\MR}{\multirow}

\begin{table}[]
    \footnotesize
    \scriptsize
    \centering
    \begin{tabular}{l@{\,}l l l | l@{\,}l l l }
    \toprule 
         \bf ISO & \bf Language & \bf  Family & \bf  Script & \bf ISO & \bf Language & \bf  Family & \bf  Script \\
    \midrule 
 af & Afrikaans & Germanic & Latin      &        ja & \textbf{Japanese} & Japonic & Kanji; Kana      \\
 da & Danish & Germanic & Latin &        ko & \textbf{Korean} & Koreanic & Hangul       \\
 nl & \textbf{Dutch} & Germanic & Latin &        vi & \textbf{Vietnamese} & Vietic & Latin \\
 de & \textbf{German} & Germanic & Latin        &        zh & \textbf{Chinese Mandarin} & Chinese & Chinese \\
 \cmidrule{5-8} \\[-10pt]
 en & \textbf{English} & Germanic & Latin       &        bn & \textbf{Bengali} & Indo-Aryan & Eastern-Nagari    \\
 is & Icelandic & Germanic & Latin      &        gu & Gujarati & Indo-Aryan & Gujarati  \\
 lb & Luxembourgish & Germanic & Latin  &        hi & \textbf{Hindi} & Indo-Aryan & Devanagari  \\
 no & Norwegian & Germanic & Latin &     kn & Kannada & Tamil & Kannada \\
 sv & \textbf{Swedish} & Germanic & Latin &      mr & Marathi & Indo-Aryan & Devanagari \\
 fy & Western Frisian & Germanic & Latin        &        ne & Nepali & Indo-Aryan & Devanagari \\
 yi & Yiddish & Germanic & Hebrew &      or & Oriya & Indo-Aryan & Odia \\
 \cmidrule{1-4} \\[-10pt]
 ast & Asturian & Romance & Latin &        pa & Panjabi & Indo-Aryan & Gurmukhi \\
 ca & Catalan & Romance & Latin &        sd & Sindhi & Indo-Aryan & \tabtl{Persian\\ Devanagari} \\
 fr & \textbf{French} & Romance & Latin &        si & Sinhala & Indo-Aryan & Sinhala \\
 gl & Galician & Romance & Latin        &        ur & Urdu & Indo-Aryan & Arabic \\
 it & Italian & Romance & Latin &        ta & \textbf{Tamil} & Dravidian & Tamil \\
 \cmidrule{5-8} \\[-10pt]
 oc & Occitan & Romance & Latin &        ceb & Cebuano & Malayo-Polyn. & Latin      \\
 pt & \textbf{Portuguese} & Romance & Latin &    ilo & Iloko & Philippine & Latin       \\
 ro & Romanian & Romance & Latin &       id & \textbf{Indonesian} & Malayo-Polyn. & Latin   \\
 es & \textbf{Spanish} & Romance & Latin        &        jv & Javanese & Malayo-Polyn. & Latin      \\
 \cmidrule{1-4} \\[-10pt]
 be & Belarusian & Slavic & Cyrillic    &        mg & Malagasy & Malayo-Polyn. & Latin \\
 bs & Bosnian & Slavic & Latin  &        ms & Malay & Malayo-Polyn. & Latin \\
 bg & Bulgarian & Slavic & Cyrillic     &        ml & Malayalam & Dravidian & Malayalam \\
 hr & Croatian & Slavic & Latin &        su & Sundanese & Malayo-Polyn. & Latin \\
 cs & \textbf{Czech} & Slavic & Latin   &        tl & Tagalog & Malayo-Polyn. & Latin \\
 \cmidrule{5-8} \\[-10pt]
 mk & Macedonian & Slavic & Cyrillic &   my & Burmese & Sino-Tibetan & Burmese \\
 pl & \textbf{Polish} & Slavic & Latin &         km & Central Khmer & Khmer & Khmer     \\
 ru & \textbf{Russian} & Slavic & Cyrillic &     lo & Lao & Kra-Dai & Thai; Lao \\
 sr & Serbian & Slavic & Cyrillic; Latin &       th & Thai & Kra-Dai & Thai \\
 sk & Slovak & Slavic & Latin &  mn & Mongolian & Mongolic & Cyrillic \\
 \cmidrule{5-8} \\[-10pt]
 sl & Slovenian & Slavic & Latin &       ar & \textbf{Arabic} & Arabic & Arabic \\
 uk & Ukrainian & Slavic & Cyrillic &    he & \textbf{Hebrew} & Semitic & Hebrew        \\
 \cmidrule{1-4} \\[-10pt]
 et & Estonian & Uralic & Latin &        ps & Pashto & Iranian & Arabic \\
 fi & \textbf{Finnish} & Uralic & Latin &        fa & \textbf{Farsi} & Iranian & Arabic \\
 \cmidrule{5-8} \\[-10pt]
 hu & \textbf{Hungarian} & Uralic & Latin       &        am & Amharic & Ethopian & Ge'ez        \\
 lv & Latvian & Baltic & Latin &        ff & Fulah & Niger-Congo & Latin       \\
 lt &  \textbf{Lithuanian} & Baltic & Latin &      ha & Hausa & Afro-Asiatic & Latin      \\
  \cmidrule{1-4} \\[-10pt]
 sq & Albanian & Albanian & Latin &      ig & Igbo & Niger-Congo & Latin        \\
 hy & Armenian & Armenian & Armenian    &        ln & Lingala & Niger-Congo & Latin \\
 ka & Georgian & Kartvelian & Georgian  &        lg & Luganda & Niger-Congo & Latin \\
 el & \textbf{Greek} & Hellenic & Greek &        nso & Northern Sotho & Niger-Congo & Latin \\
  \cmidrule{1-4} \\[-10pt]
 br & Breton & Celtic & Latin   &        so & Somali & Cushitic & Latin \\
 ga & Irish & Irish & Latin     &        sw & \textbf{Swahili} & Niger-Congo & Latin \\
 gd & Scottish Gaelic & Celtic & Latin  &        ss & Swati & Niger-Congo & Latin \\
 cy & Welsh & Celtic & Latin-Welsch     &        tn & Tswana & Niger-Congo & Latin \\
  \cmidrule{1-4} \\[-10pt]
 az & Azerbaijani & Turkic & \tabtl{Latin; Cyrillic\\ Persian}   &        wo & Wolof & Niger-Congo & Latin \\
 ba & Bashkir & Turkic & Cyrillic       &        xh & Xhosa & Niger-Congo & Latin \\
 kk & Kazakh & Turkic & Cyrillic        &        yo & Yoruba & Niger-Congo & Latin \\
 tr & \textbf{Turkish }& Turkic & Latin &        zu & Zulu & Niger-Congo & Latin \\
 \cmidrule{5-8} \\[-10pt]
 uz & Uzbek & Turkic & Latin; Cyrillic &         ht & Haitian Creole & Creole & Latin   \\
    \bottomrule 
    \end{tabular}
    \caption{\textbf{100 Languages grouped by family.} For each language, we display the ISO code, language name, language family, and script. Languages in bold are \textit{bridge languages} (\textit{Malayo-Polyn.} stands for \textit{Malayo-Polynesian}).}
    \label{tab:all_languages}
\end{table}

We consider several factors to select which languages to focus on.
First, we include widely-spoken languages from geographically diverse language families.
We cover a diversity of scripts and resource levels (as shown in Table~\ref{tab:all_languages}) to have high coverage of languages worldwide.
Second, we use languages for which public evaluation data exists, as we must be able to quantify model performance. 
Lastly, we only use languages for which monolingual data is available, as monolingual data is a critical resource for large-scale mining.
Combining these three criteria results creates our full list of 100 languages,  summarized in Table~\ref{tab:all_languages}.

\subsubsection{Evaluation Benchmarks}
\label{sec:eval_data}

We use publicly available evaluation datasets to evaluate the performance of all of our models.
To cover our set of $100$ languages and $2200$ directions, we bring together data from a variety of sources. We describe each evaluation dataset below. 

\begin{itemize}
    \item \textbf{WMT} --- The majority of language pairs from WMT go through English and the data is from the news domain. We consider data for $13$ languages~\citep{ondrej2017findings,bojar-etal-2018-findings,barrault2019findings}.
    \item \textbf{WAT} --- The WAT competition covers Asian languages paired with English. We consider data for Burmese-English~\citep{riza2016introduction}, which contains news articles. WAT contains many other evaluation directions, but many of those are covered by WMT or in a specific domain, so we focus on Burmese-English for WAT only.
    \item \textbf{IWSLT} --- The IWSLT translation competition contains data from TED talks paired with English translations. We use data for $4$ languages~\citep{cettolo2017overview}.
    \item \textbf{FLORES} --- FLORES\footnote{\url{https://github.com/facebookresearch/flores}}~\citep{flores2019} pairs two low resource languages, Sinhala and Nepali, with English in the Wikipedia domain.
    \item \textbf{TED} --- The TED Talks dataset\footnote{\url{https://github.com/neulab/word-embeddings-for-nmt}}~\citep{Ye2018WordEmbeddings} contains translations between more than $50$ languages; most of the pairs do not include English. The evaluation data is n-way parallel and contains thousands of directions.
    \item \textbf{Autshumato} --- Autshumato\footnote{\url{https://repo.sadilar.org/handle/20.500.12185/506}, CTexT® (Centre for Text Technology, North-West University), South Africa; Department of Arts and Culture, South Africa} is an $11$-way parallel dataset comprising $10$ African languages and English from the government domain. There is no standard valid/test split, so we use the first half of the dataset for valid and second half for test.
    \item \textbf{Tatoeba} --- Tatoeba Challenge\footnote{\url{https://tatoeba.org/eng/}} covers $692$ test pairs from mixed domains where sentences are contributed and translated by volunteers online. The evaluation pairs we use from Tatoeba cover 85 different languages.
\end{itemize}
  
We evaluate the quality of translations with BLEU~\citep{papineni2002bleu}. We first detokenize all data, then apply standard tokenizers for each language before computing BLEU. For most languages, we use the \texttt{moses} tokenizer~\citep{koehn2007moses}.\footnote{\url{https://github.com/moses-smt/mosesdecoder/blob/master/scripts/tokenizer/tokenizer.perl}} For Chinese we use the SacreBLEU tokenizer (\texttt{tok zh}) and convert all traditional characters generated by the model to simplified characters using HanziConv\footnote{\url{https://github.com/berniey/hanziconv}}~\citep{post2018sacrebleu},\footnote{The evaluation datasets for Chinese usually contained simplified characters. However, our training data contains a mix of simplified and traditional characters, and thus the model could generate either. We convert the generated traditional Chinese characters to simplified for consistency.} for Indian languages we use the Indic NLP library~\citep{kunchukuttan2020indicnlp},\footnote{\url{https://github.com/anoopkunchukuttan/indic_nlp_library}} for Japanese we use Kytea,\footnote{\url{https://github.com/neubig/kytea}} for Thai we use PyThaiNLP~\citep{pythainlp},\footnote{\url{https://github.com/PyThaiNLP/pythainlp}} for Arabic we use the QCRI Arabic Normalizer,\footnote{\url{http://alt.qcri.org/tools/arabic-normalizer/}} for Korean we use Mecab,\footnote{\url{https://pypi.org/project/python-mecab-ko/}} for Burmese we use the official segmentation tool provided by \citet{ding2019towards}, for Romanian we follow \citet{sennrich2016edinburgh} and apply Moses tokenization, special normalization, and remove diacritics for Romanian texts,\footnote{\url{https://github.com/rsennrich/wmt16-scripts/blob/master/preprocess/}} and finally for Serbian we transliterate the output to Latin characters before computing BLEU.\footnote{In Serbian, both Latin script and Cyrillic script are used, and often intermixed within a sentence in the evaluation data. As the target sentence could be in either script and it is not possible to predict the target script from the input, we transliterate before computing BLEU.}
We release the tokenization and evaluation scripts for reproducibility \href{https://github.com/pytorch/fairseq/tree/master/examples/m2m_100}{\texttt{here}}\footnote{\url{https://github.com/pytorch/fairseq/tree/master/examples/m2m_100}}. We remove all data from all evaluation sets from our training sets.

\subsection{Covering the Language Matrix by Mining Relevant Parallel Data}

Supervised translation systems rely on large quantities of parallel sentences, which we refer to as bitext data, which are traditionally derived from human translations.
Most existing bitext datasets go through English, with a few domain specific exceptions such as proceedings from international organizations~\citep{Koehn:2005:mtsummit_eurparl,ziemski:2016:un_corpus}.
These corpora are limited in size and domain, and an alternative is to \textit{mine} parallel data~\citep{resnik1999mining,utiyama2003reliable} in large collections of monolingual data~\citep{conneau2019unsupervised,wenzek2019ccnet}.
In this work, we leverage and extend the corpus provided by two of these mining projects: CCMatrix~\citep{schwenk2019ccmatrix} and CCAligned\footnote{\url{http://www.statmt.org/cc-aligned}}~\citep{elkishky2020ccaligned}.
In the following, we describe our mining strategy and summarize the main ideas of CCMatrix and CCAligned. We refer the reader to the references for a detailed description of the approaches.

\paragraph{Mining parallel data with LASER.} 
Mining parallel data consists of searching for sentences that could be potential translations in large monolingual corpora.
This search requires a measure that captures the semantic similarity between sentences in different languages.
Most recent methods build this similarity by comparing the embeddings from a neural network trained on multilingual data~\citep{artetxe2018margin,chen:2020:acl_mine,bojar:2020:acl_mine}.
We focus on the embeddings generated by the LASER encoder, which enables the comparison of sentences in  $94$ different languages~\citep{Artetxe:2018:arxiv_massive_ml}. We then retrieve parallel corpora efficiently using FAISS indexing~\citep{johnson2019billion}.
LASER embeddings generalize to unseen languages, like Asturian, allowing us to mine bitexts for $100$ languages.
The generic data mining pipeline consists of several steps: \textbf{(1)} a large corpus of text is preprocessed and divided into different languages, \textbf{(2)} candidate pairs of aligned sentences are embedded and stored in a index, \textbf{(3)}  indexed sentences are compared to form potential pairs, \textbf{(4)} the resulting candidate pairs are filtered in post-processing. 

\paragraph{CCMatrix Mining Approach.}
CCMatrix takes a global approach: all unique sentences in one language are compared with all unique sentences in another language.
This \textit{global mining} approach has the advantage of considering all possible documents when searching for the translation of a sentence.
CCMatrix works on the large monolingual corpora in the $91$ languages of CCNet~\citep{wenzek2019ccnet}, but at this scale, the global search is computationally demanding even with fast indexing from FAISS~\citep{johnson2019billion}.
Thus, we apply it to a selected subset of relevant pairs, as detailed in \autoref{sect:bridge}.

\paragraph{CCAligned Mining Approach.}
CCAligned avoids the scaling challenges of global mining by pre-selecting documents to compare.
This \textit{local mining} follows a hierarchical approach: first, document-level language identification along with various rules is applied to find whole documents that are likely to contain mutual translations~\citep{elkishky2020ccaligned}. Parallel sentences are then mined using LASER-based alignment within the paired documents only. Filtering~\citep{chaudhary2019low} is performed to remove unaligned data that exists  because the original webpage did not have any parallel data, only partial parallel data, or other processing failures.
One advantage of this approach is that it is very fast, scalable, and retrieves parallel sentences with high precision. Another is that each English document is aligned to many non-English documents --- thus, mining non-English pairs can be quickly performed by joining non-English documents paired to the same source.

\paragraph{Postprocessing.}
We apply a filtering step to remove sentences of greater than 50\% punctuation. 
The data is then deduplicated, and we remove any sentence that appears in any validation or test dataset -- even if it is associated with another language pair.
Finally, we apply length and language-specific filtering.
The length filtering removes sentences that are too long -- more than $250$ subwords after segmentation with SPM -- or with a length mismatch between the sentence and its translation -- if the length ratio is greater than $3\times$.
The language-specific filtering removes sentences that contain more than $50\%$ of characters that have not been marked as core to the identified language -- specifically, characters that are commonly used in the identified language with the exception of white space, numbers, punctuation, and Latin characters for non-Latin script languages.


\subsubsection{Bridge Language Group Mining Strategy}
\label{sect:bridge}

\begin{figure}[t]
    \centering
    \includegraphics[width=0.9\textwidth]{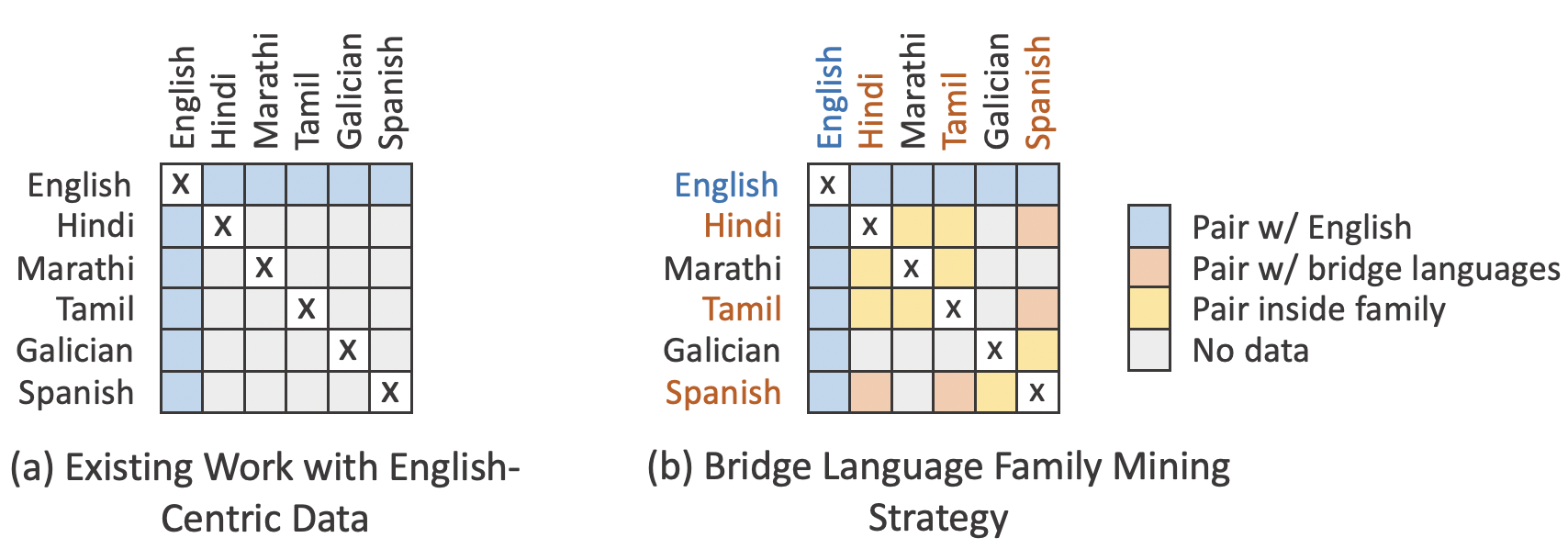} 
    \caption{\textbf{Depiction of an English-Only data mining setting compared to the Bridge Language Mining Strategy}. We display a data matrix, where languages are shown on the X and Y axes. Data is mined in one direction (such as Hindi to Marathi) and used to train bidirectionally.\\
    }
    \label{fig:data_fig0}
\end{figure}

Mining data for each and every language pair is prohibitive --- previous work circumvents this issue by focusing only on the $99$ pairs that go through English~\citep{zhang2020improving}. One alternative to the extensive computation required to mine all possible combinations of pairs is \textit{sparse mining}, or mining only a select subset of pairs. A straightforward strategy is to \textit{randomly} select pairs to mine, but this does not use any linguistic information on how languages are related and spoken around the world.

In this work, we propose an alternative based on language families and bridge languages that avoids exhaustively mining every possible pair.
Our goal is to reduce the number of bitext pairs while preserving translation directions of practical interest.
We first group all the $100$ languages into $14$ \textit{language groupings}.
All languages within a grouping are mined against each other.
For instance, within the Indic language grouping, we mine all pairs of Bengali, Hindi, Marathi, Tamil, Urdu, and so on.
 The motivation for this strategy is two-fold.
First, people living in areas that speak multiple languages in the same grouping tend to communicate a lot with each other and would benefit from high quality direct  translation.
Second, systematically mining languages of the same grouping is helpful for training language-specific parameter models (see \autoref{sec:lang_specific}).

For the most part, languages are grouped by linguistic similarity, e.g. Germanic, Slavic, or Malayo-Polynesian languages. 
However, the size of the resulting groupings varies greatly, resulting in less mined data for the languages in the smallest groupings.
We further group languages by geographic and cultural proximity to reduce this discrepancy.
For example, Uralic and Baltic languages are gathered into a single group to increase the quantity of mined data.
The resulting groupings are shown in Table~\ref{tab:all_languages}.

To connect languages across groupings, we define 1--3 \textit{bridge languages} in each grouping, usually those with the most resources, such as
Bengali, Hindi, and Tamil for the $12$ languages in the Indo-Aryan family.
All $26$ bridge languages are highlighted in Table~\ref{tab:all_languages}.
These bridge languages are mined against all other bridge languages.
Finally, all $100$ languages are mined against English.
We illustrate this mining strategy in Figure~\ref{fig:data_fig0}.
On the left, we depict what many current approaches model: data only through English.
On the right, we depict our Many-to-Many language matrix for several example languages.
Compared to English-Centric, our dataset has far greater coverage of non-English, direct translation directions.

\begin{figure}[t]
    \centering
    \includegraphics[width=\textwidth]{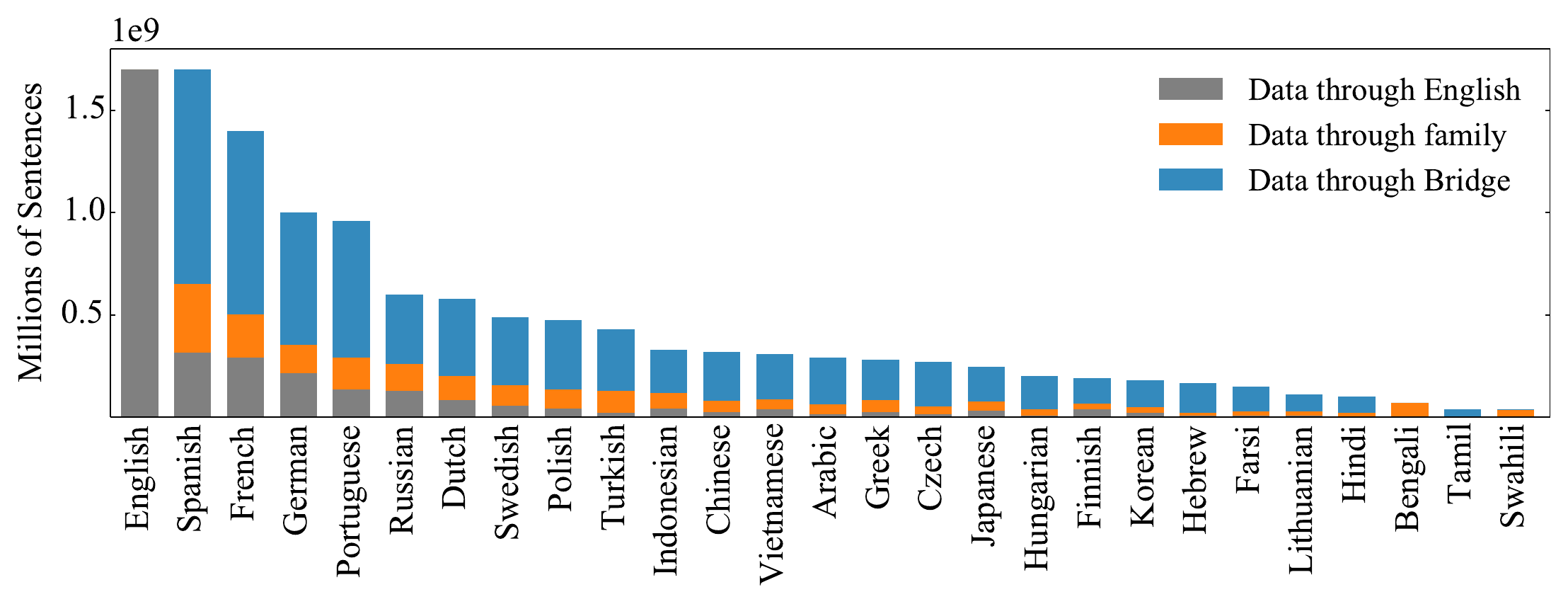} 
    \caption{\textbf{Total Amount of Data through Bridge Languages on our 100x100 Training Corpus}. We depict the amount of data through English (gray), amount of data through a bridge language not counting English (orange), and amount of data through the language grouping not counting bridge languages (blue).}
    \label{fig:data_statistics}
\end{figure}

\paragraph{Training Data Statistics.}
In total, our final training dataset contains 7.5B parallel sentences, corresponding to $2200$ directions.
In Figure~\ref{fig:data_statistics}, we show all bridge languages and demonstrate how their associated training data is divided between translations with English, within a language grouping, or with bridge languages across language groupings.
Of particular interest is the comparison between the additional Many-to-Many data and the data through English.
We observe that 5--10 times more parallel data can be mined if using a Many-to-Many strategy, compared to an English-Centric one. This is particularly beneficial for mid- and low-resource languages.

\subsubsection{Results} 

\begin{table}[t]
    \begin{minipage}{0.44\textwidth}
    \setlength{\tabcolsep}{5.5pt}
    \centering
    \small 
        \begin{tabular}{l | c | c c c }
        \toprule
        & \bf All & \multicolumn{3}{c}{\bf  Supervised} \\ 
        \bf Model& \bf  Avg & \bf  Low &  \bf  Mid & \bf  High   \\
        \midrule 
        Random 80\% & 11.9 & 3.6 & 16.1 & 31.5  \\  
        Random 80\% w/ En & 16.3 & 8.9 & 22.4 & 36.6  \\  
        \midrule
        Bridge Language, 80\% & \bf 17.2 & \bf 10.4 & \bf 23.2 & \bf 37.4  \\  
        \bottomrule
        \end{tabular}
    \end{minipage}
    \hfill 
    \begin{minipage}{0.44\textwidth}
        \centering 
        \includegraphics[width=\linewidth]{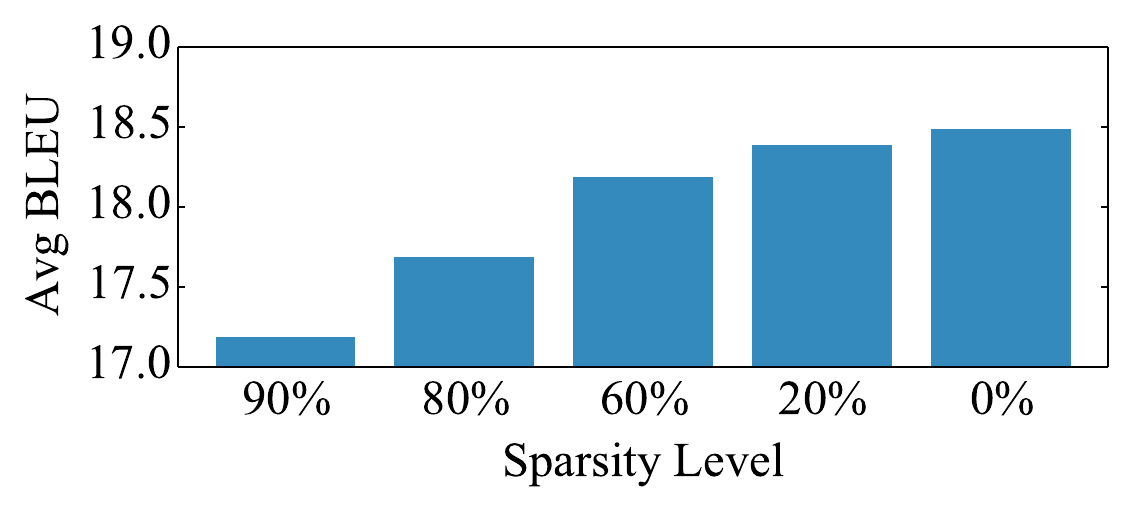}
    \end{minipage}
    \caption{\textbf{(left) Comparison of Sparse Mining Strategies}. We first hold the sparsity level fixed at 80\% --- compared to randomly selecting pairs to mine, the Bridge Language mining strategy performs very well. \textbf{(right) Bridge Language Strategy at Different Sparsity Levels}. We analyze different levels of sparsity in the language matrix to understand how many pairs to mine. Based on these results, we take 60\% sparse as a tradeoff point between strong performance and reasonable quantity of mining. 0\% indicates no sparsity, or a fully mined language matrix.} 
    \label{tab:mining_strategy1}
\end{table}

We validate the impact of several decisions made during data construction.
First, we study the impact of our bridge language strategy compared to English-Centric mining augmented by other random pairs, as well as fully random mining.
Second, we investigate the impact of the level of sparsity chosen in our bridge strategy, focusing on a subset of 50 languages.

\paragraph{Bridge Language strategy versus Random and English-Centric Random.}
We experimentally evaluate the impact of our bridge language mining strategy on the performance of our baseline model in Table~\ref{tab:mining_strategy1} (left). 
We consider two additional baselines, a fully random mining strategy (Random 20\%) and a \textit{English-Centric + Random} strategy (Random 20\% w/ En).
In the Random strategy, mined pairs are randomly chosen, while in the \textit{English-Centric + Random} strategy, we retain all pairs through English and only select the remaining pairs randomly.
We show that fully random mining has a substantial negative impact on performance, as a lot of high quality data is aligned through English, so sampling fully randomly eliminates a large portion of the potential training set.
Random 20\% w/ En is worse as well. Through examination, we find that randomly sampling pairs to mine often selects pairs that do not produce as much data, as the pairs may not include high resource languages. However, the bridge language strategy ensures that high resource languages are mined, and then focuses on mining languages in related families. This produces a large amount of bitext, and at the same time, covers many language directions.

\paragraph{Impact of Sparsity.}
We control the sparsity of our language matrix using the number of bridge languages. In Figure~\ref{tab:mining_strategy1} (right), we show the impact of sparsity on the performance of our baseline model compared to a fully mined language matrix (0\% sparse).
We observe that increasing the amount of mined data to make the matrix less sparse is helpful, but fully mining the matrix is not substantially better.
The main reason is that our mining strategy prioritizes frequently used pairs which are often associated with the largest bitext, while the discarded pairs are often associated with small bitext. For example, fully mining the matrix would mine a pair such as Icelandic to Chinese, but the amount of data produced by mining this pair is quite low. This case is representative of what occurs as the full matrix is mined --- as increasingly more data is mined, the additional pairs begin to add less data which in turn leads to diminishing quality improvements.

%
%

\subsection{Augmenting Bitext Data with Backtranslation}
\label{sec:bt}

Backtranslation (BT) creates synthetic bitexts from unaligned monolingual data~\citep{Schwenk:2008:unsup,bojar2011bt_pbmt,sennrich2016backtranslation,edunov2018bt,hoang2018iterative}.
The core idea is to translate monolingual sentences in the backward direction, and add the obtained synthetic translations to the training set.
More precisely, when training a model to translate from a source language to a target language, backtranslation generates additional data by translating monolingual target sentences into the source language.
Using backtranslation thus requires the ability to translate in both directions, which fits well into the setting of multilingual machine translation ~\citep{zhang2020improving,siddhant2020leveraging}.
However, generating these backtranslations is time consuming even for a single direction, which is compounded in the Many-to-Many case.
We thus focus on applying backtranslation on specific pairs to supplement mining data where needed.

\paragraph{Selecting Backtranslation Pairs.} 
Our goal is to translate between 100 languages and to provide good translation quality for as many translation directions as possible. 
To this end, we use BT to improve directions which have initially lower translation quality.
We identify these language directions by measuring the quality of our 1.2B parameter multilingual model before applying BT.
Since back-translation is computationally intensive, we focus on $100$ directions with a BLEU score of between $2$ and $10$.
For $50$ of these directions, we do not have any bitext at all as we did not mine all 4,450 possible language pairs.

\begin{figure}[t]
    \centering
    \includegraphics[width=\textwidth]{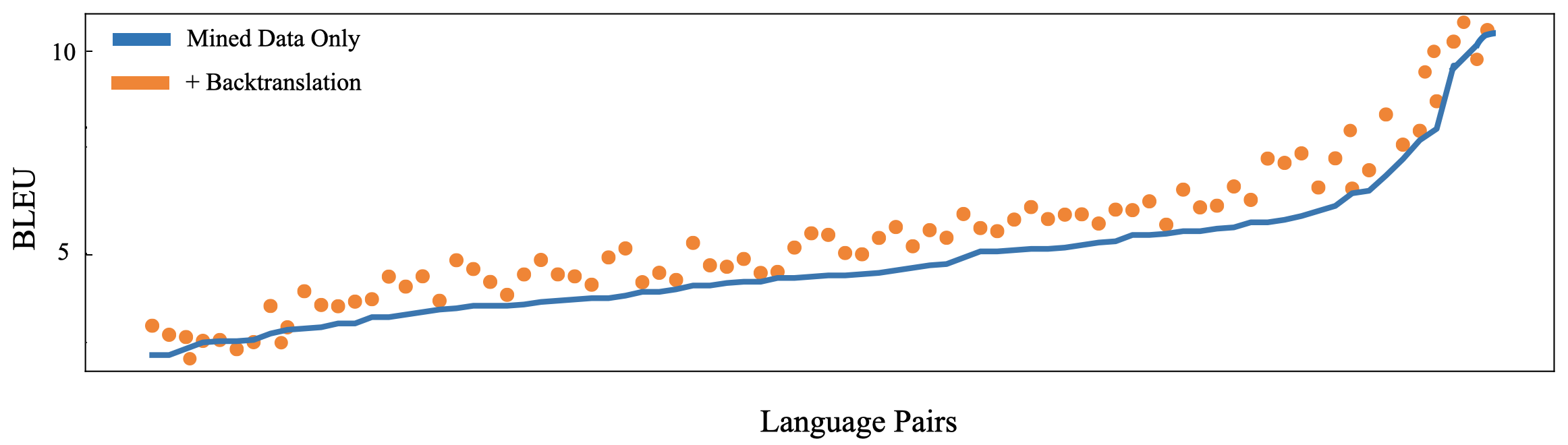} 
    \caption{\textbf{Improvements from Adding Backtranslated Data.} For each of the 100 language directions we explored by adding backtranslation. The blue line indicates the original model, where directions were selected if they had between 2 and 10 BLEU. The orange scatter indicates the effect of adding backtranslation. Languages are ordered by their original BLEU scores before backtranslation.}
    \label{fig:bt_fig1}
\end{figure}

\paragraph{Training a Multilingual Model with Additional Backtranslations.}
For the selected pairs, we first generate synthetic translations that are added to the training set without upsampling.
Following~\citet{chelba2019tagged}, we add a special encoder-side BT token to these translations to indicate to the model that they are synthetic.
For each of the $100$ target languages, we randomly sample $50$ million unique monolingual sentences from the cleaned CommonCrawl corpus of \citet{wenzek2019ccnet}.
The synthetic translations are then generated with our $1.2$B MMT model.
We use a beam search with beam of size $5$ and fix all the hyper-parameters, including the length penalty, to the same values for all the directions.
We apply the same filtering to the backtranslations as the original mined training data, which substantially reduces the size of the resulting synthetic bitexts.

\paragraph{Impact of Backtranslated Data.} 
Results are shown in Figure~\ref{fig:bt_fig1}, where we compare the original Many-to-Many model used to create the backtranslations (blue line) with the improvements after training a multilingual model with the backtranslation added (orange scatter).
Backtranslation almost always improves performance for any direction, regardless of the original BLEU score.
As the amount of data generated with BT correlates with the length of training time, we decide to focus on applying BT on directions with low performance (BLEU between 2 and 10) to improve our MMT system where it underperforms.

%
%

\subsection{Balancing Languages in a Many-to-Many Setting}

\begin{table}[t]
\setlength{\tabcolsep}{5.5pt}
\centering
\small 
    \begin{tabular}{l  c  cccc c c c  c}
    \toprule
    &~~&  \multicolumn{4}{c}{\bf  Supervised} && \bf  Zero-Shot && \bf All\\ 
\cmidrule{3-6}
    \bf Data sampling     	&& \bf  Low &  \bf  Mid & \bf  High & \bf  Avg && \bf  Avg && \bf  Avg\\
    \midrule 
    Uniform               	&& 6.1 & 20.4 & 38.4 & 19.0  && 11.8 			&& 15.7     \\  
    Temperature Rescaling 	&& 10.2 & 23.7 & 38.0 & 21.8 && 13.0 			&& 18.1     \\  
    \midrule                                                                                       
    Sinkhorn Temp. Rescaling    && \bf 10.9 & \bf 24.1 & \bf 38.3 & \bf 22.2 && \bf 13.5 && \bf 18.6 \\  
    \bottomrule
    \end{tabular}
    \caption{\textbf{Comparison of Various Sampling Strategies.} 
We report BLEU on the validation set of our $1.2$B base multilingual model trained with different data sampling schemes.
Performance is broken down into different resource-setups (low, mid, high) where bitext data exists (supervised) or in the zero-shot setting for pairs without data.} 
    \label{tab:temp_ablation}
\end{table}

The data distribution produced by large-scale mining is not balanced between languages, so training a model would favor over-represented languages.
A standard solution is to rebalance each language independently with Temperature Sampling~\citep{arivazhagan2019massively}, e.g. replacing the probability $p_\ell$ of a language by $p_\ell^{\frac{1}{T}}$. 
In the English-centric case, changing the probability of sampling a language changes the probability of the other languages only through the normalization.
However, in the Many-to-Many case, language distributions are more interdependent.
For example, some languages are only paired with a subset of other languages or to an overrepresented language.
Thus, sampling them will affect the probability of these other languages they are paired with.
This strategy thus has no guarantee to produce the target balanced distribution between languages.
We describe \textit{Sinkhorn Temperature Sampling}, which extends the temperature sampling strategy to the Many-to-Many setting.  

Our goal is to design a sampling technique such that the distribution of languages on the source \textit{and} target sides is equal to a given target distribution.
Unfortunately, sequentially sampling the source language and then the target would not work, as some languages are only paired with a subset of languages --- making it impossible to sample the target language according to a given distribution.
Moreover, the sizes and distributions of bitexts greatly vary from a language to another.
Instead, we propose directly sampling a pair of languages from a matrix of pair probabilities such that the marginal distributions of languages corresponds to our target distribution.
In practice, this means that each row and column of the matrix should sum to the probability of the corresponding language.
More precisely, we estimate a square matrix $\mathbf{P}^*$ such that:
$$\max_\mathbf{P} ~\text{tr}~ \left( \mathbf{P Q} \right) ~~~~~~\text{s.t.}~~~\mathbf{P}1_L = \mathbf{p}^{\frac{1}{T}},~\mathbf{P}^\top1_L = \mathbf{p}^{\frac{1}{T}},$$
where $\mathbf{p}$ is the vector stacking the probabilities of the $L$ languages and $\mathbf{Q}$ is the matrix of pair probabilities.
This problem can be solved exactly with the Sinkhorn-Knopp algorithm.
The matrix $\mathbf{Q}$ has entries equal to $0$ for pairs with no bitext and this algorithm preserves them in the solution $\mathbf{P}^*$,
hence adding no probability mass to missing bitexts. 
We calculate this once before training and set the temperature $T$ to $5$.
In Table~\ref{tab:temp_ablation}, we show the benefits of this strategy over temperature sampling with a constant improvement of $0.5$ in BLEU.

%% file: comp.tex
\section{Many-to-Many Compared to English Centric}
\label{sec:comp}

In this section, we first present an experiment to better understand the performance improvements of English-Centric systems and to compare them to our Many-to-Many setting.

\paragraph{Experimental Setting.}
We train our $1.2$B model on the full $100$ language Many-to-Many dataset and compare it to the same model trained only on data through English.
We use the same vocabulary built with SentencePiece on the full dataset in both cases.
Each model has a different dataset size and we train for 500K updates.
This number of updates corresponds to one pass over the entire Many-to-Many dataset and $3.5$ passes on the English-centric data.
We tune the dropout rate for each model over the values $\{0.1, 0.2, 0.3\}$.

\subsection{Main Result}

\begin{table}[t]
    \setlength{\tabcolsep}{5.5pt}
    \centering
    \small 
        \begin{tabular}{l  ccc}
        \toprule
        \bf Setting & \bf To English & \bf From English & \bf Non-English \\
        \midrule 
        Bilingual baselines & 27.9 & \bf 24.5 & 8.3 \\ 
        English-Centric & 31.0 & 24.2 & 5.7  \\ 
        English-Centric with Pivot & --- & --- & 10.4 \\
    \midrule
        Many-to-Many & \bf 31.2 & 24.1 & \bf 15.9 \\    
        \bottomrule
        \end{tabular}
    \caption{\textbf{Comparison of Many-to-Many and English-Centric Systems.} Many-to-Many matches the performance of English-centric on evaluation directions involving English, but is significantly better on non English directions.
    } 
    \label{tab:m2m_english}
\end{table}

\begin{table}[t]
    \begin{minipage}{0.47\textwidth}
        \centering
        \small 
            \begin{tabular}{l   cc}
            \toprule
            \bf Setting &  \bf w/ bitext & \bf w/o bitext\\
            \midrule 
            En-Centric & 5.4 & 7.6 \\ 
            En-Centric Piv. & 9.8 & 12.4 \\
            \midrule
            M2M  & \bf 12.3 & \bf 18.5 \\    
            \bottomrule
            \end{tabular}
        \caption{
\textbf{Many-to-Many versus English-Centric on zero-shot directions.}
We report performance on language pairs with and without bitext in the Many-to-Many training dataset.
	} 
        \label{tab:m2m_english2}
    \end{minipage}
    \hfill 
    \begin{minipage}{0.47\textwidth}
        \centering
        \small 
            \begin{tabular}{l  ccc}
            \toprule
            \bf Setting & \bf $\rightarrow$En & \bf En$\rightarrow$ & \bf Non-En \\
            \midrule 
            En-Centric & \bf 26.4 & 17.8 & 2.4  \\ 
            En-Centric Piv. & --- & --- & 5.1 \\
            \midrule
            M2M & 25.7 & \bf 18.1 & \bf 9.4 \\    
            \bottomrule
            \end{tabular}
        \caption{
\textbf{Many-to-Many versus English-Centric on one pass of data.} 
We report performance for models after a number of updates equivalent to the size of the English-centric dataset.
        } 
        \label{tab:m2m_english_onepass}
    \end{minipage}
\end{table}

In Table~\ref{tab:m2m_english}, we compare the performance of both models on different types of directions, namely, any language to English (To English), English to any language (From English), and all the directions not involving English (Non-English).
Performance is aggregated over $150$ directions for To English and From English, and over 2500 directions for Non-English.
On the pairs including English, both models achieve similar performance, suggesting that a $1.2$B model does not underfit even though the additional non-English data represents $98\%$ of the directions and 74\% of the data. 
For the non-English pairs, we consider two translation strategies for the English-Centric model: directly translating as if the model was trained on the pair -- by using the corresponding language tokens -- or by pivoting through English.
Our model outperforms direct translation with the English-Centric model by $10.2$ BLEU and when the English-Centric model uses pivoting by $5.5$ BLEU.
While this result is not surprising, it confirms that a purely English-Centric model has limited potential on non-English pairs, and there is a fundamental need for training on Many-to-Many data. 

\subsection{Understanding the Source of Improvement}
 
The main impact of adding Many-to-Many data is on the directions that do not include English.
In this section, we provide a detailed study of where we observe the largest improvements with the additional data.

\paragraph{Impact on Zero-shot.}
Many non-English pairs are not covered by our Many-to-Many model, and we can thus study if the improvements we observe originate primarily from directions associated with bitext data or if we observe the same improvement on directions where the Many-to-Many model generates translations in a zero-shot fashion.
In Table~\ref{tab:m2m_english2}, we show the performance if the evaluation is split between the Non-English pairs \textit{with} and \textit{without} bitext.
On directions with bitext, the Many-to-Many model outperforms the English-Centric model by $7$ BLEU for direct translation, and by $3.5$ BLEU for English-Centric with pivoting.
This shows the importance of diverse data.
Not surprisingly, this gap is even bigger on pairs without bitext. 
Many-to-Many  performs nearly $11$ BLEU better than the English-Centric model for direct translation, and with pivoting the gain remains over $6$ BLEU.

\paragraph{Impact of the quantity of training data.}
A hypothesis to explain the gain between English-Centric and Many-to-Many models is the effect of additional source and target side training data.
Even if the Many-to-Many system has never seen a direction at training time, it benefits from additional source and target side data available through other training pairs.
As mining non-English language pairs creates more training data compared to English-centric datasets, the Many-to-Many model benefits from a larger training set.
In Table~\ref{tab:m2m_english_onepass}, we compare both models after seeing the same quantity of data.
We train both models for one epoch. The English-Centric model performs better on To English directions, likely because it only has one output language to learn, but the Many-to-Many model outperforms on From English directions and Non-English directions.

\begin{figure}[t]
\centering
\includegraphics[width=0.9\textwidth]{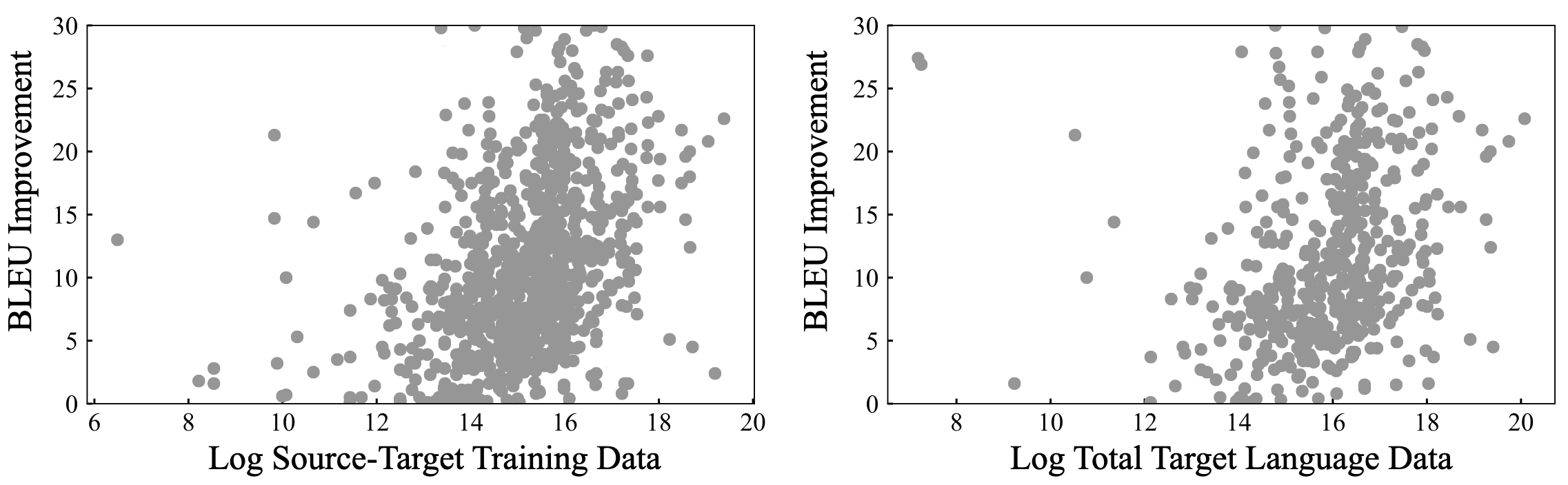} 
\caption{\textbf{Improvement of Many-to-Many over English-centric} with respect to the amount of mined training data (left) and the amount of target side language data (right). Improvement from a Many-to-Many model correlates with greater amounts of bilingual training data with Pearson correlation 0.38 (left) and greater amounts of target language data with Pearson correlation 0.32 (right).
}
\label{fig:bt_fig3}
\end{figure}

\paragraph{Which Pairs Improve the Most?}
The main factor for improvement is the quantity of data associated with either a pair or a language.
Pairs that have a large quantity of mined data, such as Spanish-Portuguese, greatly benefit from our Many-to-Many dataset.
We show this effect in the left panel of Figure~\ref{fig:bt_fig3} (left).
A second source of improvement is observed on languages for which the Many-to-Many dataset contains a large amount of data across many pairs.
This data benefits the decoder-side language model in a way that is comparable with BT.
In the right panel of Figure~\ref{fig:bt_fig3}, we show the impact of this cumulative monolingual data on the average performance per language.
Finally, we also observe a third type of improvements from the similarity in vocabulary and syntax from related languages.
A striking example is the quality of translation between English and Belarusian, where the Many-to-Many model achieves 12.7 BLEU on the TED evaluation set, compared to 3.2 BLEU for a bilingual model.
The number of bitexts for Belarusian is small, but Belarusian is related to Russian, and  the Many-to-Many model transfers its knowledge from Russian to Belarusian.

\begin{figure}[t]
\centering
\includegraphics[width=\textwidth]{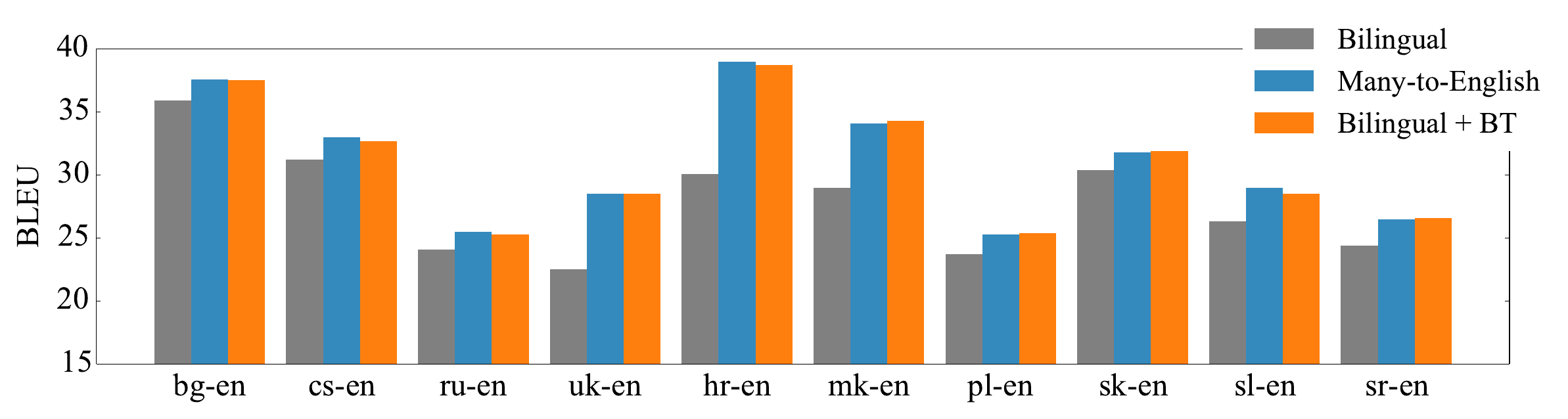} 
\caption{\textbf{Performance of many-to-English multilingual translation compared to bilingual baselines trained on mined data and bilingual + backtranslation.}
The average performance of many-to-English is 25.1 BLEU compared to 25.2 BLEU for back-translation while the bilingual system achieves 23.1.
}
\label{fig:bt_fig2}
\end{figure}

\subsection{Understanding the Performance of English-Centric Systems}

In Table~\ref{tab:m2m_english}, we confirm an observation made in~\citet{arivazhagan2019massively} that an English-Centric model improves the most over bilingual models on the directions into English, while improvement in the other directions (From English) remain more modest. 
A hypothesis to explain this discrepancy between directions from and to English is that the decoder of an English-Centric model learns a better English language model by leveraging the aggregated English data across all through-English directions.

\paragraph{Result.} 
We test this hypothesis with a controlled experiment where we compare a Many-to-English model with bilingual models using backtranslated English data (\autoref{sec:bt}).
The experiment is based on 11 Slavic languages and we backtranslate the exact same English data as was used to train the Many-to-English model so that both models are trained on the same English data. 
Figure~\ref{fig:bt_fig2} shows that backtranslation performs comparably to the Many-to-English approach. 
While this improves our understanding of Many-to-English translation, a multilingual approach nevertheless retains the advantage of combining many directions into a single model which greatly simplifies modeling.

%% file: model.tex
\section{Components for Scaling Multilingual Translation Models}
\label{sec:model}

Our goal is to build a single model capable of translating $9,900$ language directions covering $100$ languages.
This creates several challenges for models with insufficient capacity to capture that many languages and scripts adequately.
To this end, previous MMT work has considered different types of large capacity models~\citep{arivazhagan2019massively,lepikhin2020gshard}.
In this section, we investigate different ways to add capacity to an MMT model:
we first investigate dense scaling, where we increase the depth and width of standard Transformer architectures.
Then, we identify disadvantages of dense scaling, and propose an alternative to effectively add \textit{language-specific} parameters and exploit the nature of language similarities within the task of multilingual machine translation.

\subsection{Dense Scaling}

\subsubsection{Background: Model Parallel Training}

During the training of a neural network, we need to fit its weights, activations, gradients, and optimizer state in memory.
This restricts the maximum capacity of a network that we can train on a single accelerated device such as a GPU.
In this section, we describe two directions to circumvent this limitation.
The first direction focuses on fitting a larger model on single device by reducing the memory required by activations and optimizer states during the training process. The second direction focuses on efficient training of even larger models through model parallelism e.g. splitting a model across multiple devices. 
In this work, we pursue both techniques to densely scale the capacity of Transformers.

\paragraph{Reducing Memory Consumption on a GPU.}  To reduce the amount of memory, we consider optimizer state sharding and gradient checkpointing.
Optimizer state sharding~\citep{rajbhandari2019zero} divides the optimizer state across distributed data parallel workers so that each worker only needs to store a fraction of the optimizer state.
We also apply gradient checkpointing, which saves memory by discarding intermediate activations before the forward pass finishes~\citep{chen2016grad}.
During the backward pass, these activations are recomputed again as required.
This trades time for memory.
In the case of a Transformer based architecture, applying gradient checkpointing at pipeline parallel model partition boundaries reduces the memory used by activations by almost 50\%. 

\paragraph{Models Sharded across Multiple GPUs.} 
Reducing the memory consumption enables fitting greater model capacity on a single GPU, but the physical limitations of a single device still apply.
A solution is to split the model into separate components that are dispatched across different GPUs and trained in parallel.
This type of solution scales model capacity with the number of GPUs.
There are two broad paradigms to split a model: along the width or along the depth.
Tensor parallelism~\citep{shoeybi2019megatron,shazeer2018mesh} splits by width, while pipeline parallelism~\citep{huang2019gpipe,kim2020torchgpipe} splits by depth, placing different layers on different GPUs. 
We use pipeline parallelism, but both methods work equally well with Transformers.
We use the implementation from \texttt{fairscale}\footnote{\url{https://github.com/facebookresearch/fairscale}}.

\begin{table}[t]
  \centering
  \includegraphics[width=0.98\linewidth]{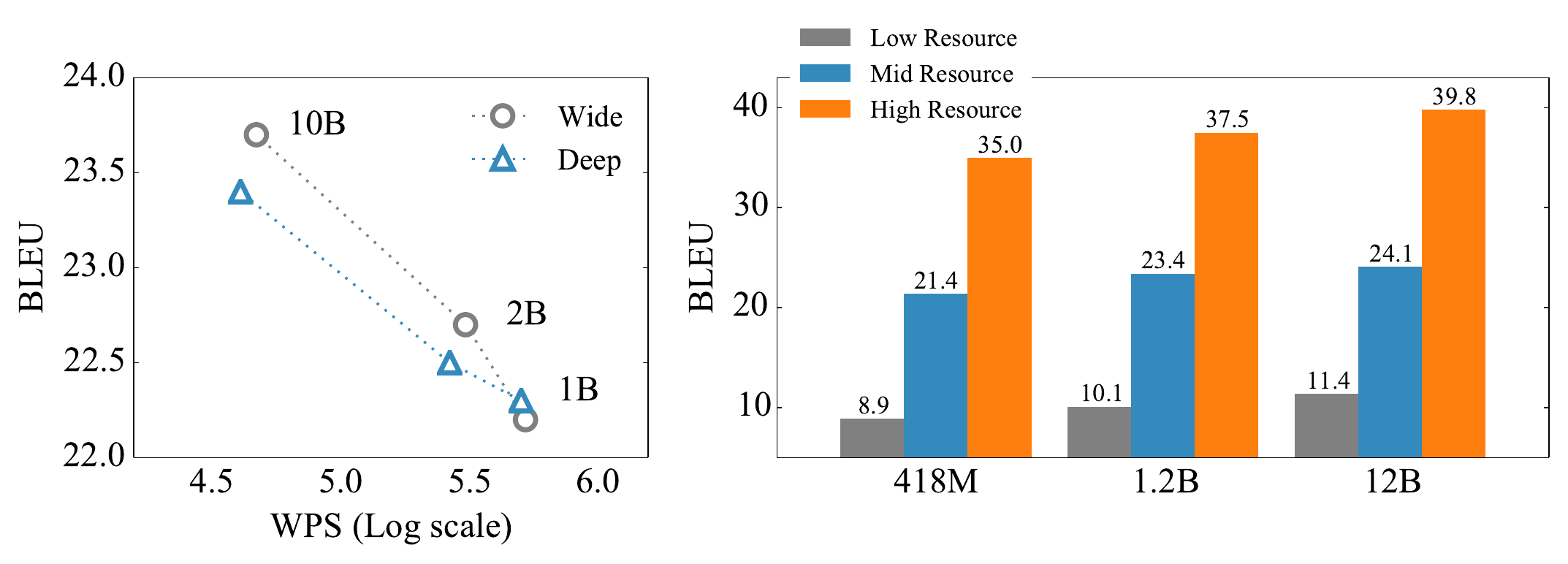}
    \captionof{figure}{
\textbf{(left) Comparison between deep versus wide models.} 
We compare the performance in BLEU for different wide and deep models as a function of their words per second (WPS) at training time, evaluating on a subset of 38 directions.
\textbf{(right) Performance of wide models for different parameter sizes.}
We compare the performance of different wide models on different pairs at low, mid, and high resource levels, evaluating on all supervised evaluation pairs.
The white lines indicate comparisons between the different models at the same resource level.
    \label{fig:dense}
    }
\end{table}

\subsubsection{Training Large Dense Models}

We investigate several strategies to increase the capacity of a sequence-to-sequence Transformer model in the context of multilingual machine translation.

\paragraph{How to Scale: Wide or Deep?} 
We consider increasing the capacity of a Transformer by either increasing the number of layers (depth axis) or the dimensions of each layer, including the feedforward (width axis).
On the left panel of Figure~\ref{fig:dense}, we analyze which axis to prioritize by comparing models with different sizes, $1$B, $2$B, and $10$B, obtained by growing their depth or width (see Appendix~\ref{app:dense} for model configurations and dimensions).
We report their performance in BLEU and their inference speed measured in words per second (WPS).
We train these models on a dataset that covers $80$ languages and evaluate them on $38$ different benchmark directions with more than $1$k parallel sentences per direction.
The main result of this study is that wider models scale better than deeper models in terms of performance and WPS.
In the rest of this paper, we thus focus on wider models.

\paragraph{Performance as a function of scale.}  
In the right panel of Figure~\ref{fig:dense}, we compare the performance of wide models as we increase their capacity from $418$M to $12$B parameters.
We train these models on the full set of $100$ languages and evaluate them on all supervised evaluation pairs.
We report their performance in BLEU for pairs with either low, mid or high resource training data.
First, as we increase the number of parameters, we observe that the performance increases, \textit{even on low-resource pairs}.
This suggest that even a $12$B parameter model could be underfitting with our many-to-many multilingual dataset.
However, improvements increase roughly logarithmically in the number of parameters, and we need to scale model size by an order of magnitude to improve by a few BLEU points, e.g., $+1.5$ BLEU from $1.2$B to $12$B.
As we scale models densely, their runtime and memory usage becomes too prohibitive to justify the gain in performance, and so, we consider alternatives to increase the capacity of our models more efficiently.

\subsection{Scaling Model Capacity with Language-Specific Parameters}
\label{sec:lang_specific}

In this section, we introduce a layer whose parameters are split by language or language group based on similarity in vocabulary.
Each translation direction only accesses a subset of these parameters, allowing the model capacity to scale without significantly affecting the training and inference time.
The layer is trained with a novel re-routing scheme to improve generalization which we detail below. Compared to previous work~\citep{wang-etal-2018-three,bapna2019simple,zhang2020improving}, we focus on allocating entire language-specific layers and using this to scale model size while maintaining training speed.

\begin{figure}[t]
    \centering 
    \includegraphics[width=0.9\textwidth]{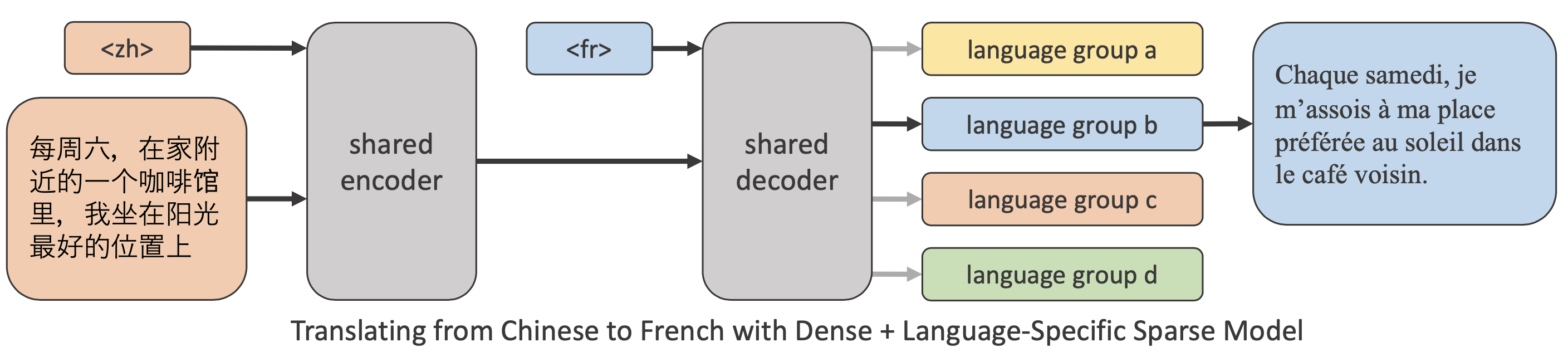} 
    \caption{\textbf{Language-Specific Parameters} provide specialized capacity to an otherwise fully shared multilingual encoder and decoder.}
    \label{fig:langspec_fig0}
\end{figure}

\paragraph{Parallel Transformer Layer.}
We follow the sequence-to-sequence Transformer architecture and replace some of its layers by a set of parallel Transformer layers, one for each pre-defined group of languages. 
More precisely, assuming we have split the languages into $K$ fixed groups, this parallel layer is composed of $K$ parallel Transformer sublayers, one per language group.
For each translation, we then select the corresponding sublayer among the $K$ possibilites depending on the language direction.
If the parallel layer is in the encoder, we select the sublayer according to the source language, while if it is in the decoder, we select according to the target language.
In practice, we only add these layers to either the encoder or decoder, not both.
This enables us to split translations along with their sublayers per GPU, leading to faster training and efficient memory usage.
Figure~\ref{fig:langspec_fig0} shows an example of the resulting \textit{trunk-and-branch} architecture when the parallel layer is in the decoder.

\paragraph{Grouping Languages by Frequency and Similarity.} 
We group languages based on two criteria: the amount of training data and their vocabulary.
The motivation for these criteria is that we can learn a specific layer for a language with enough data, and for the rest, overlapping vocabulary is a good proxy for similar languages.
First, each language with more than $100$M sentences forms its own group and hence has its own sublayer.
We have $28$ languages that fit this criteria: hu, ru, hi, ro, fr, nl, fi, pt, ar, el, vi, en, ms, tr, he, id, pl, cs, sv, fa, zh, bg, de, es, ko, ja, it, da.
Second, we group the remaining languages by vocabulary overlap, leading to $18$ additional groups. To create these groupings, we calculate the vocabulary overlap between the training data of different languages and cluster those that have high overlap together. Note that some low resource languages have their own script --- such as Kannada --- and are not clustered with any similar languages as the script is unique. However, to maintain balance between groups~\citep{wang2020balancing}, we cluster the remaining languages together and roughly balance the amount of training data for each group. 
In total, we form $46$ groups, each with its own sublayer in a language-specific layer.

\paragraph{Random Re-Routing between Sublayers.}    
During training and inference, a sublayer is deterministically selected according to its language direction.
This guarantees that our model always uses the same memory and time during inference, regardless of the translation pair.
However, during training, this deterministic routing does not share information between similar languages if not associated with the same sublayer.
For example, the sublayer associated with Ukrainian does not benefit from the large quantity of Russian training data, since Russian has its own isolated sublayer.
We mitigate this shortcoming by \textit{random re-routing} of translations, i.e., randomly picking another sublayer instead of the designated one. 
This shares information between languages associated with different sublayers,  benefiting low resource languages by training on similar high resource languages. The re-routing is completely random, though could be restricted to re-route only to similar languages.

\paragraph{Adding Language-Specific layers to Pre-Trained Transformers.} 
We can integrate a language-specific layer into an already pre-trained Transformer by adding it either at the end of the decoder or at the beginning of the encoder.
We can then freeze the parameters of the pre-trained Transformer and learn the language-specific components. These additional language-specific layers train rapidly as the rest of the model already has strong performance. 
This strategy means it is straightforward to adapt pre-trained networks to a new domain or language by training a small number of dedicated parallel layers, and could easily be extended to various other applications.

\subsubsection{Evaluation of the Language-Specific Layer}

We experiment with different scenarios by adding a language-specific layer to the encoder or decoder, or to a pre-trained densely scaled model. We demonstrate the importance of random re-routing.
Finally, we validate this strategy by comparing it to scaling models densely.

\paragraph{Parallel layer in Encoder or Decoder?}
The trunk-and-branch architecture for language-specific layers is general and can be used to specialize capacity for any neural architecture. We explore adding language-specific capacity in the encoder or decoder using a smaller setting of 10 high-resource languages. Table~\ref{tab:langspec_scaling_encdec} shows that language-specific parameters are generally more effective when applied to the decoder network. 
Recent studies show that encoders are more important for bilingual machine translation~\citep{wu2019pay,kasai2020deep}, however, these studies are based on systems modeling only a single language direction compared to our setting. 
In our case, increasing the encoder or the decoder does not impact performance significantly, and we focus on decoder for the rest of this paper.

\begin{table}[t]
    \begin{minipage}{0.47\textwidth}
    \setlength{\tabcolsep}{5.5pt}
    \centering
    \small 
        \begin{tabular}{lc  c}
        \toprule
        \bf Model & \bf Params & \bf BLEU\\
        \midrule 
        Language Specific Enc & 540M & 17.1 \\ 
            & 920M & 17.5 \\ 
        \midrule 
        Language Specific Dec & 540M & 17.3 \\ 
            & 920M & 17.8 \\ 
        \bottomrule
        \end{tabular}
        \caption{
\textbf{Comparing Language-Specific Encoders and Decoders}. 
We add parallel language-specific layers to either the encoder or decoder, with different sizes.
} 
        \label{tab:langspec_scaling_encdec}
    \end{minipage}
    \hfill 
    \begin{minipage}{0.5\textwidth}
        \vspace{-0.15cm}
        \centering 
        \includegraphics[width=\linewidth]{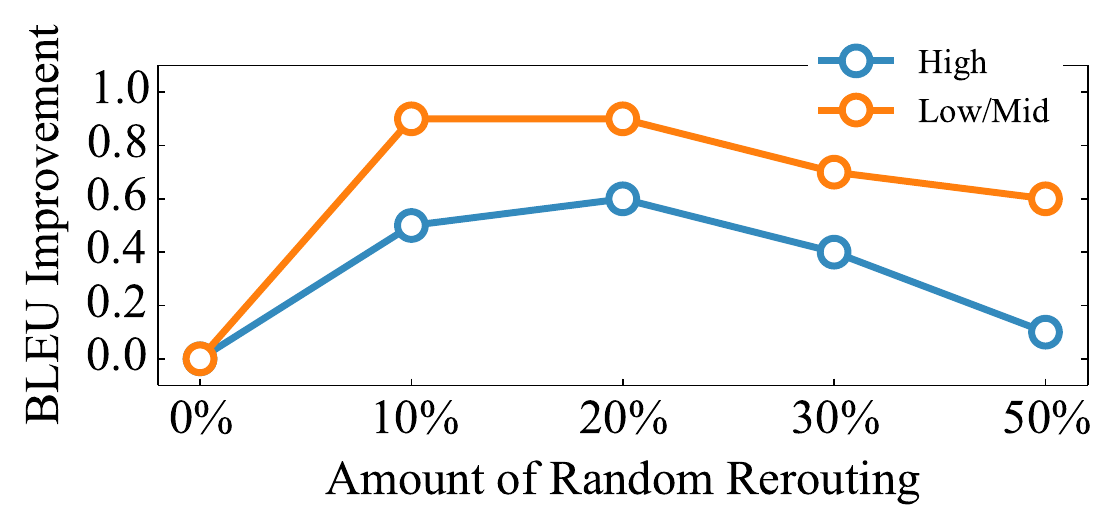}
       \captionof{figure}{\textbf{Impact of Re-routing Rate} on the performance of high resource and low/mid resource languages.}  
        \label{tab:rerouting}
    \end{minipage}
\end{table}

\paragraph{Random Re-routing.}
Figure~\ref{tab:rerouting} shows the impact of the re-routing strategy on performance as we increase the number of training samples routed to random parallel layers as opposed to their assigned layers.
With a re-routing rate of 20\%, an improvement of about 0.8 BLEU can be achieved over no re-routing for low and mid resource languages, without affecting the performance of high resource languages.
Too much stochasticity leads to performance similar to no random re-routing for high resource languages, but still improves mid to low resource performance compared to no re-routing. 

\begin{table}[t]
\setlength{\tabcolsep}{5.5pt}
\centering
\small 
    \begin{tabular}{lll  c  ccc c c  c}
    \toprule
    & & &~~& \multicolumn{3}{c}{\bf  Supervised} && \bf All\\ 
\cmidrule{5-7}
    \bf Model & \bf Params & \bf WPS	&& \bf  Low &  \bf  Mid & \bf  High && \bf  Avg\\
    \midrule 
    Dense Transformer & 1.2B & 40K && 10.1 & 23.4 & 37.5 && 17.5 \\ 
    Dense Transformer  & 3B & 20K && 10.3 & 23.8 & 38.0 && 17.9 \\ 
    Dense Transformer  & 12B & 3.5K && \bf 11.8 & 24.2 & 39.9 && 18.6 \\ 
    \midrule 
    Dense Transformer 1.2B \\ 
    ~~~with 1 Language-Specific Layer & 1.9B & 38K && 10.7 & 24.1 & 38.5 && 18.1 \\ 
    ~~~with 3 Language-Specific Layers & 3.5B & 34K && 10.6 & \bf 24.7 & 39.5 && 18.8 \\ 
    ~~~with 6 Language-Specific Layers & 10B & 26K && 10.5 & 24.7 & \bf 40.3 && \bf 19.2 \\ 
    \bottomrule
    \end{tabular}
    \caption{\textbf{Scaling Model Size with Language-Specific Parameters}. 
We start with a $1.2$B parameter baseline with $24$ encoder layers and $24$ decoder layers.
We add increasingly more decoder layers to language specific layers.
For example, in the case of 1 language-specific decoder layer, the decoder has 23 shared layers and 1 language-specific layer.
We demonstrate the effect of using 1, 3, and 6 language specific layers. The additional parameters for language-specific layers are split across all language groups. We report WPS at training time holding the batch size fixed on 8 GPUs. The 12B baseline uses model parallel.
    } 
    \label{tab:langspec_scaling}
\end{table}

\paragraph{Comparison with Large Dense Models.} 
We compare adding language specific capacity with densely scaling model size in Table~\ref{tab:langspec_scaling} on $100$ languages.
As language-specific layers add many parameters, we compare to baseline models at various sizes for multiple points of comparison.
Our conclusion is that language-specific layers improve results compared to baselines of similar parameter size, particularly for mid and high resource languages where there is sufficient data to train the language-specific sublayers. Further, compared to dense scaling, sparse scaling only uses a fraction of the parameters in each forward pass, which maintains fast training speed despite large total model size.

\begin{figure}[t]
    \centering 
    \includegraphics[width=\textwidth]{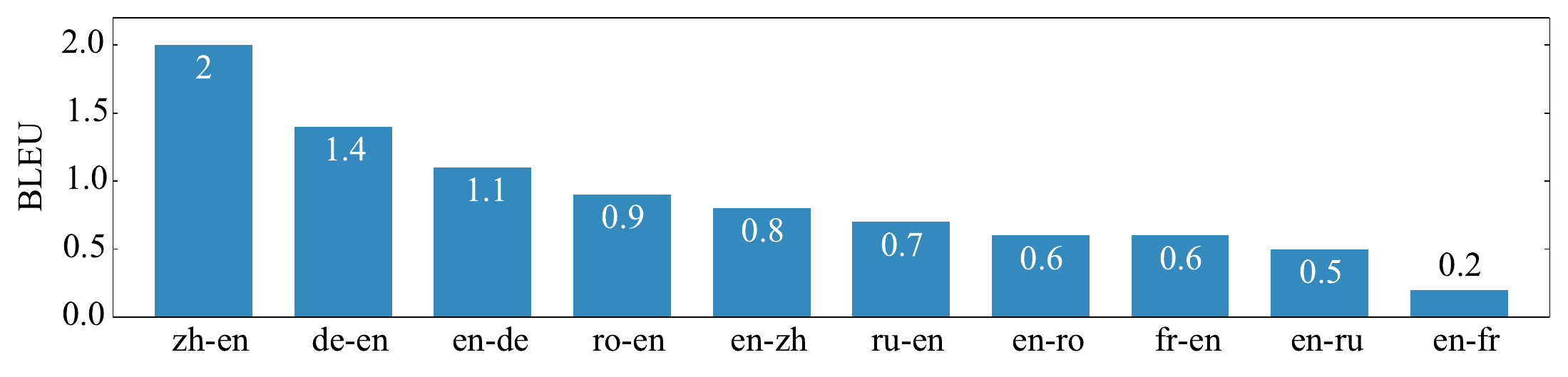} 
    \caption{\textbf{BLEU Improvement using Dense + Sparse} over Dense alone. We display evaluation on a subset of pairs using the WMT evaluation datasets.}
    \label{fig:dense_sparse}
\end{figure}

\paragraph{Adding Language-Specific Layers to a Pre-Trained Model.} 
We demonstrate the impact of adding language-specific layers to the decoder of a pre-trained $12$B parameter Transformer in Figure~\ref{fig:dense_sparse}. We show that adding language-specific layers for five languages improves results on the WMT evaluation datasets.
The language-specific layer adds $3.4$B parameters and we train it for $20$K updates with the rest of the network frozen. The total size of this model is $15.4$B parameters.
For several directions, we observe gains of more than 1 BLEU, which validates this strategy. On average, we observe gains of 0.9 BLEU. 

%% file: results.tex

    



%% file: appendix.tex

\section{Additional Information about Data} 
\label{app:dictionary}

\begin{figure}[h!]
    \centering 
    \includegraphics[width=\textwidth]{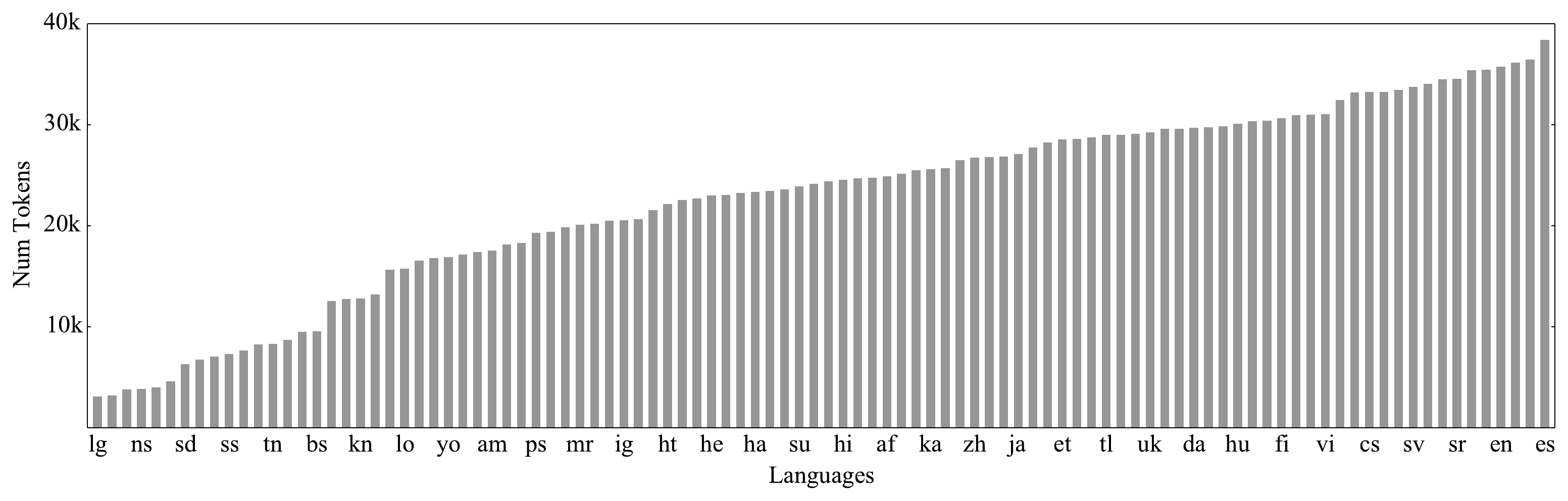} 
    \caption{\textbf{Dictionary Coverage} per Language}
    \label{fig:tok_per_lang}
\end{figure}

Figure~\ref{fig:tok_per_lang} displays the dictionary coverage for each of our 100 languages. 

\section{Model Architectures}
\label{app:dense}

Table~\ref{tbl:wide_deep} shows the various model configurations considered in our experiments when scaling dense models. 

\begin{figure}[h!]
    \centering
        \begin{tabular}{l c c c }
        \toprule
        \bf Size &  \bf Embed & \bf FFN & \bf Layers \\ 
        \midrule
        1B Wide & 1024 & 16K & 14 \\ 
        1B Deep & 1024 & 4K & 38 \\ 
        \midrule
        2B Wide & 2048 & 16K & 11 \\ 
        2B Deep & 1024 & 8K & 48 \\ 
        \midrule
        10B Wide & 4096 & 16K & 24 \\ 
        10B Deep & 3072 & 12K & 36 \\ 
        \bottomrule
        \end{tabular}
    \caption{\textbf{Architecture of Wide and Deep Models.}
    }
    \label{tbl:wide_deep}
\end{figure}

\section{Exploiting Multilinguality at Inference Time with Multi-source Self-Ensembles}

Throughout the paper, we explored how to improve the performance of single models, scaling the amount of data as well as the model size, but there remain numerous directions for future investigation of multilinguality. One direction is understanding how to exploit the nature of multilingual translation at inference time as well. 

A known, effective strategy to improve accuracy is to ensemble multiple models at inference time. 
However, this requires training multiple models which substantially increases the training compute requirements.
Instead, we suggest exploring self-ensembles, created by applying the multilingual model to the same source sentence in different languages.
For example, if we wish to translate Galician to English, then instead of directly translating between the two, we ensemble the translation of Spanish to English with the translation of Galician to English, using the same multilingual model for both directions, and by averaging the predicted token log-probabilities, as for standard multi-model ensembles. 
The additional source is obtained by translating the input to another \textit{intermediary} language.
After this, we ensemble the translation of both sources to the target.
This uses the same multilingual model for all steps.

\begin{table}
  \centering
  \small
  \begin{tabular}{lr}
    \toprule 
    \bf Model & \bf BLEU  \\
    \midrule 
    Multilingual & 17.3 \\
    Multi-Model Ensemble & 17.5 \\
    Pivoting with Multilingual & 17.0 \\ 
    Multi-source Self-Ensemble & 17.5 \\ 
    \bottomrule 
  \end{tabular}
  \caption{\textbf{Results on zero-shot language pairs for Multi-Source Self-Ensemble} compared to various baselines. We report the average test BLEU score on 100 randomly sampled pairs.}
  \label{tab:multisource_zeroshot_results}
\end{table}

We evaluate both pivoting and self-ensembling on zero-shot directions as these can benefit from better accuracy. 
We report results on 100 randomly sampled zero-shot translation directions which have at least 1000 examples in the validation and test set.
Next, for each translation direction, we choose the intermediary language that resulted in the highest BLEU on the validation set; the same is done to choose the intermediary language for pivoting.
We also tune a weight to balance the two language directions~\citep{garmash2016ensemble}.
Table~\ref{tab:multisource_zeroshot_results} shows that multi-source self-ensembling improves the single model result by 0.2 BLEU on average.
It also performs as well as standard multi-model ensembling but requires training only a single model.
This is particularly relevant for large models trained on vast quantities of data, which require a lot of compute to be able to perform standard ensembling.